\documentclass[10pt,twocolumn,twoside]{IEEEtran}

\usepackage{blindtext}
\usepackage{graphicx}
\usepackage{multirow}
\usepackage[table]{xcolor}
\usepackage{collcell}
\usepackage{hhline}
\usepackage{pgf}
\usepackage{rotating}
\usepackage{subcaption}
\usepackage{amsmath}
\usepackage{ amssymb }
\usepackage{tikz}
\usetikzlibrary{shapes}
\usetikzlibrary{arrows}
\usetikzlibrary{decorations.pathmorphing}
\usepackage{lipsum}
\usepackage{algorithm}
\usepackage[noend]{algpseudocode}
\usepackage{url}
\usepackage[noadjust]{cite}
 \usepackage{multirow}
\usepackage{bm}
\newcommand\gray{gray}

\newcommand\ColCell[1]{%
  \pgfmathparse{#1<0.85?1:0}%
    \ifnum\pgfmathresult=0\relax\color{white}\fi
  \pgfmathparse{1-#1}%
  \expandafter\cellcolor\expandafter[%
    \expandafter\gray\expandafter]\expandafter{\pgfmathresult}#1}

\newcolumntype{E}{>{\collectcell\ColCell}c<{\endcollectcell}}

\makeatletter
\newcommand*\titleheader[1]{\gdef\@titleheader{#1}}
\AtBeginDocument{%
  \let\st@red@title\@title
  \def\@title{%
    \bgroup\normalfont\large\centering\@titleheader\par\egroup
    \vskip1.5em\st@red@title}
}
\makeatother

\titleheader{This work has been submitted to the IEEE for possible publication. Copyright may be transferred without notice, after which this version may no longer be accessible}
\title{Person Identification using Seismic Signals generated from Footfalls }

\begin{document}


%

\author{Bodhibrata\,Mukhopadhyay,
        Sahil\,Anchal
        and~Subrat\,Kar
\thanks{B. Mukhopadhyay, S. Anchal and S. Kar  are with the Department
of Electrical Engineering, Indian Institute of Technology Delhi,
Hauz Khas,New Delhi, 110016 India (e-mail: bodhibrata@gmail.com, sahilanchal1021@gmail.com, subrat@ee.iitd.ac.in).
}

}


\maketitle

\begin{abstract}
\boldmath 
Footfall based biometric system is perhaps the only person identification technique which does not hinder the natural movement of an individual. This is a clear edge over all other biometric systems which require a formidable amount of human intervention and encroach upon an individual's privacy to some extent or the other. This paper presents a Fog computing architecture for implementing footfall based biometric system using widespread geographically distributed geophones (vibration sensor). Results were stored in an Internet of Things (IoT) cloud. We have tested our biometric system on an indigenous database (created by us) containing 46000 footfall events from 8 individuals and achieved an accuracy of 73\%, 90\% and 95\% in case of 1, 5 and 10 footsteps per sample. We also proposed a basis pursuit based data compression technique \textit{DS8BP} for wireless transmission of footfall events to the Fog. \textit{DS8BP} compresses the original footfall events (sampled at 8 kHz) by a factor of 108 and also acts as a smoothing filter. These experimental results depict the high viability of our technique in the realm of person identification and access control systems.

\end{abstract}


\IEEEpeerreviewmaketitle

\section{Introduction} \label{Sec:intro}

\IEEEPARstart{S}{ensors} like Passive Infrared (PIR)~\cite{fang2006path}, camera~\cite{wang2003silhouette}, fingerprint scanner~\cite{jain2000biometric}, accelerometer~\cite{Begg_svm_gait}, interferometric reflectance imaging sensor (IRIS)~\cite{iris_lie}, microphone~\cite{Wang}, mm wave scanner~\cite{body_shape_mm_wave_person}, pressure sensor fabric~\cite{carpet_Zhou}etc. have been extensively used in biometric systems. However, most of these popular biometric sensors need active cooperation of the individual. Some require them to stand in front of a camera while others need them to place their finger on a scanner or keep an eye on the IRIS. However, in footfall based biometric system, the individuals are only required to walk through the active region of the sensor. The main advantages of this type of biometric system are: seismic sensors can be easily camouflaged; evading detection is impossible because footstep patterns are inimitable; it does not breach individual's privacy; less sensitive to environmental parameters and beyond the capacity of an individual to decode and manufacture the raw signal. Human identification system has significant applications in various areas such as office premises, classrooms, shopping malls, military areas, buildings, hospitals etc. It can be used for registering attendance of students in a classroom, employees in an office or workers in a workshop just by placing a seismic sensor or an array of sensors at the entrance gate. It has the capability to detect intruders in high security areas by identifying footsteps of unregistered users. Applications can also be developed to control home appliances in smart buildings using footfall~\cite{jun2015controlling}.

This paper implements a human identification system using the signatures of the seismic waves generated from footfalls. The seismic vibration created by rhythmic movement of heel and toe during walking is captured using a geophone. Processing a seismic signal in the ``pay-as-you-go" cloud in real time requires high computational power and bandwidth. So, in our implementation, we shifted the computational functions to the edge of the network using Fog computing paradigm~\cite{Preden_fog_mist,fog_computing_Tang}. 

The main contributions of the paper are as follows: i) It proposes and implements a Fog based architecture for person identification using seismic waves generated by footfalls. ii) A footstep database containing a total of 46000 footfall events from 8 individuals were created.
iii) It proposes a basis pursuit based data compression technique \textit{DS8BP} for reducing bandwidth and energy during wireless transmission of the raw seismic signal. iv) Experiments have been conducted to obtain the best suited classifier (both in terms of accuracy and computational complexity), minimum sampling frequency of the signal and number of footsteps required per sample to achieve a respectable accuracy. v) Studies have also been carried out to find the minimum number of footsteps required (for each individual) to train a classifier for achieving a certain classification accuracy. vi) Experimental validation of the proposed model \textit{DS8BP} have been done and compared with other techniques.

The remainder of the paper is organized as follows: In Section~\ref{Sec:rel_work} we discuss the related works on footfall based human identification system. In Section~\ref{Sec:sys_arch} we introduce the Fog based paradigm for implementing an IoT based person identification system. Section~\ref{Sec:result_diss} presents the performance analysis of the proposed techniques. Finally, conclusion and future works are discussed in Section~\ref{Sec:future_con}.

\begin{table*}[]
\centering
\caption{Comparison among existing footstep based biometric system.}
\label{Tb:survey}
\resizebox{\textwidth}{!}{\begin{tabular}{|l|l|l|l|l|l|l|l|}
\hline
\multirow{2}{*}{Author(s)} & \multirow{2}{*}{Year} & \multirow{2}{*}{Organ} & \multirow{2}{*}{Sensor} & \multicolumn{2}{c|}{Database} & \multirow{2}{*}{ $\text{Algorithm}^\text{\textdagger}$} & \multirow{2}{*}{$\text{Accuracy}^\ast$} \\ \cline{5-6}
 &  &  &  & \# Person(s) & \# Samples &  &  \\ \hline
Addlesee \textit{et al.} \cite{Addlesee} & 1997 & Weight & Load cell  & 15 & 300 & HMM & IA: $<$50\% \\ \hline
Middleton \textit{et al.} \cite{middleton_floor} & 2005 & Foot pressure & Pressure sensor  & 15 & 180 & Decision trees & IA: 80\% \\ \hline
Suutala \textit{et al.} \cite{SUUTALA200821} & 2007 & Foot pressure & Electromechanical film   & 11 & 440 & SVM & IA: 93.96\% \\ \hline
Qian \textit{et al.} \cite{Qian} & 2010 & Footfall & Pressure sensing floor  & 11 & 5690 & LDA & IA: 92\% \\ \hline
    Miyoshi \textit{et al.} \cite{Miyoshi_microphone} & 2011 & Footfall & Microphone  & 12 & 720 & GMM & IA: 92.8\% \\ \hline
  Yun \textit{et al.} \cite{Yun} & 2011 & Footfall & Photo sensor  & 10 & 5000 & MLP & IA: 96.64\% \\ \hline
V.Rodriguez \textit{et al.} \cite{Rodriguez_footstep_mat} & 2013 & Foot pressure & Piezoelectric sensor & 120 & 20,000 & BTime, BSpace & EER: (5-15)\% \\ \hline
%
 %
\multirow{3}{*}{Wang \textit{et al.} \cite{Wang}} & \multirow{3}{*}{2015} & \multirow{3}{*}{Walking interval} & Mic (carpet platform) & 15 & 450 & \multirow{3}{*}{Similarity based} & Pre: 91.78\% \\ \cline{4-6} \cline{8-8} 
 &  &  & Mic (carpet platform) & 15 & 600 &  & Pre: 87.50\% \\ \cline{4-6} \cline{8-8} 
 &  &  & Mic (wooden platform) & 15 & 450 &  & Pre: 86.68\% \\ \hline
 %
 %
Pan \textit{et al} \cite{Pan_Indoor_Person_Identification} & 2015 & Footfall & Seismic & 5 & - & SVM & IA: 93\% \\ \hline
Serra \textit{et al.} \cite{serra} & 2016 & Foot pressure & Polymer floor sensor & 13 & 529 & SVM & IA: 76.9\% \\ \hline
 %
 Pan \textit{et al}\cite{Pan_2017} & 2017 & Footfall & Seismic & 10 &  & SVM & IA:96\% \\ \hline
 %
 %
Zohu \textit{et al.} \cite{carpet_Zhou} & 2017 & Foot pressure & Fabric based pressure sensor & 13 & 529 & SVM & IA: 76.9\% \\ \hline
 %
%
Proposed & 2017 &  Footfall &  Seismic & 8 & 46,489 & SVM & IA: 90\%, 95\% $^{\ast \ast}$ \\ \hline
\multicolumn{8}{|l|}{\begin{tabular}[c]{@{}l@{}}
\textdagger  Hidden Markov Model (HMM); Decision trees (DT); Linear Discrement analysis (LDA); Gaussian Mixture Model (GMM); Multiple Layer Perceptron (MLP); \\ Hamming Distance (HD); Support Vector Machine (SVM); Artificial Neural Network (ANN); Linear Regression (LR) 
\\ $\ast$ Identification Accuracy (IA); (EER); Precision (Pre); False Acceptance Rate (FAR); False Rejection Rate (FRR)
\\ $\ast \ast$ 90\% (5 footstep/sample); 95\% (10 footstep/sample)

\end{tabular}} \\ \hline
\end{tabular}}
\end{table*}

\section{Related Works}  \label{Sec:rel_work}

Broadly biometric properties can be divided into two categories, behavioural (gait, keystroke, signature, voice, footsteps) and physiological (face, fingerprint, palmprint, signature, DNA etc)~\cite{Wang,jain_biometric}. A biometric system~\cite{fog_face} collects biometric data from a person, extracts useful information from it, compares the extracted features with a standard database and predicts the identity of the person using pattern recognition techniques. Details of the existing biometric systems are presented chronologically in Table~\ref{Tb:survey}. 

Yun \textit{et al.} \cite{Yun} proposed a system for human identification called UbiFloorII, which consisted of uniformly distributed photo interrupters that converted reflected light to voltage. They developed software modules to extract gait patterns of users from footfall samples. Their system consisted of 24 square tiles of length 30 cm and each tile consisted of 64 photo sensors (sensor density = 700 per $m^2$). V.Rodriguez \textit{et al.} \cite{Rodriguez_footstep_mat} used spatiotemporal information of the footstep signals collected using piezoelectric sensor for person recognition. They observed that the identification accuracies remain almost same (5\% to 15\% EER) in case of temporal and spatial approaches. However, the accuracy increases if the temporal and spatial features are fused together. They used two $45 \times 30 \ cm^2$ mats each containing 88 piezoelectric sensors (sensor density - 650 per $m^2$).  Zohu \textit{et al.} \cite{carpet_Zhou} used a flexible fabric based pressure sensor for person identification and achieved an accuracy of 76.9\% with 13 participants. The fabric consisted of parallel electrodes on top and bottom layer separated by a non-conducting polymer substrate. The mat produced 120-by-54 pressure points and had a sensing area of 1.8\,m by 0.8\,m. Thus, person identification using footfall pressure based techniques require high number of sensors and have a small sensing region. On the contrary, in case of seismic signal based techniques, only a single geophone is required and they have a circular sensing range of radius 2.5 m(approx) in concrete. So, it is easier to scale up a system using seismic signal based approach.

DeLoney \textit{et al.} \, \cite{deloney2008person} proposed a person identification system by classifying sounds generated by footsteps. They used spectro-temporal modulations for classification. They achieved an accuracy more than 60\% for identifying 9 people wearing three different types of shoes on two types of floors. Miyoshi \textit{et al.} \cite{Miyoshi_microphone} used a microphone based technique for person identification. They recorded footstep data, detected footfalls, extracted features like Mel-Frequency Cepstral Coefficients (MFCCs), $\Delta$MFCCs, and $\Delta$Logarithm Powers, and used k-Nearest Neighbour (k-NN) and GMM to predict the identity of the person. Wang \textit{et al.} \cite{Wang} also proposed a microphone based person identification system. They collected datasets from 15 examinees in three different types of scenarios: examinees walking on concrete passage wearing their favorite shoe, examinees walking on a concrete passage wearing 4 types of shoes, and examinees walking on a wooden passage. Guo and Wang \cite{guo_mic_iet} used a Bayesian decision classifier to predict identities of individuals using features from footstep data recorded with the help of a microphone. The main drawback of using a microphone is that it gets affected by environmental noise and audible sounds.

Pan \textit{et al.} \,\cite{Pan_Indoor_Person_Identification} were the first research group to use structural vibration for human being identification. They extracted features related to gait patterns from different individuals. They calculated prediction accuracy using features from a single footstep and also from consecutive footsteps. They considered five footsteps with high SNR (Signal to Noise ratio) for prediction and achieved an average accuracy of 83\%. They have also performed confidence level thresholding which increased their accuracy to 96.5\%. Confidence level thresholding eliminated potential incorrect classification cases (it discarded almost 50\% of them) and tagged them as non-classifiable. 
            
\begin{figure*}[]
\centering
\includegraphics[width=\linewidth]{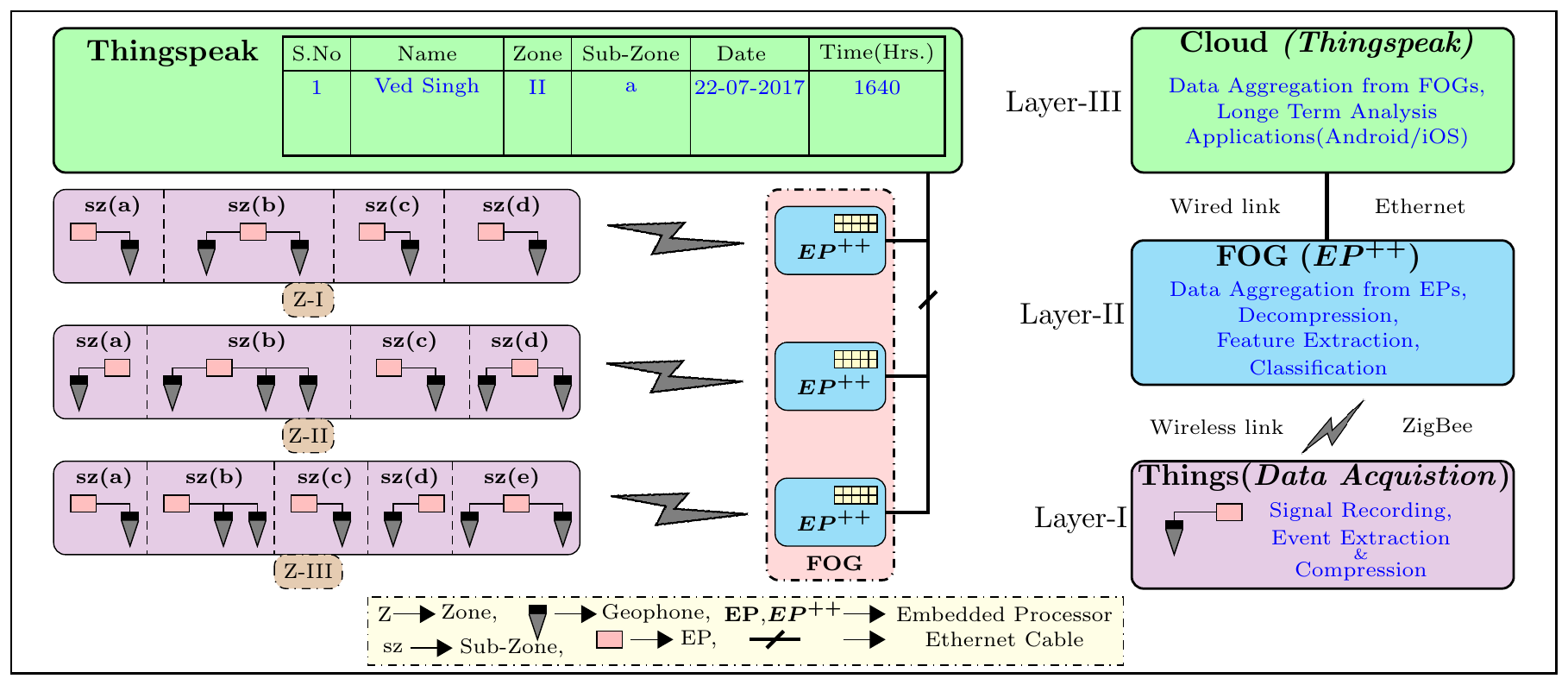}
\caption{Three layer Fog computing architecture for person identification using seismic waves generate from footfalls. }
\label{fig:sysarch}
\end{figure*}
\section{System Architecture for footstep based Person Identification} \label{Sec:sys_arch}

In this paper, we present a Fog computing based architecture for implementing human identification system using seismic sensors (refer Fig.~\ref{fig:sysarch}). The proposed system is developed to monitor people moving around different regions in real time. Depending on applications, the entire monitoring area (housing complexes, colleges, factories, hospitals etc.) is divided into zones like buildings (for housing complex), departments(for college), factory floors (for factory), hospital blocks (for hospitals) etc. and sub-zones like room (of buildings), classrooms (of departments), workshop sheds (of factory floors), and hospital wards (of hospital blocks) etc. Each sub-zone consists of a single or multiple sensors responsible for identifying humans. A huge amount of bandwidth and energy will be required to transmit the entire seismic data to the cloud (for processing), as these sensors generate enormous amount of data. To overcome this problem we shifted the computational parts from the IoT cloud~\cite{iot_survey} to the edge of the network. Commercially available clouds (ThinkSpeak, Google Cloud, Microsoft Azure etc.) are costly, bear high latency and require huge communication bandwidth. So, it is not possible to implement a cloud based system which is connected to a large number of geographically distributed bulk data generating sensors. Hence for implementing the system we propose a three layer architecture as shown in Fig.~\ref{fig:sysarch}: i) \textit{Things} (Sensor + Low end Processor: $EP$ + Transceiver)  ii) \textit{Fog} ( Embedded Processor: $EP^{++}$ + Transceiver) iii) \textit{Cloud} (Server). The proposed hierarchical structure diminishes the communication bandwidth and reduces the burden of the cloud by parallelizing the computational processes.       

\textit{Things}: In each of the \textit{Thing}, a lower end embedded processor($EP$ -- Raspberry Pi Zero/ Orange Pi Zero) is connected to seismic sensor(s) and a low power long range transceiver module (XBee 868LP). The data transmission rate of the transceiver nodes are usually in the range of 80-100~Kbps (data rate of XBee 868LP is 80 kbps) which makes it impractical to transmit each sample of the seismic signal recorded by $EP$ to the \textit{Fog}. So, $EP$ uses an event extraction technique to find the portion of signal that represents a footfall and then compresses the event before transmitting. Each sub-zone has a single \textit{Thing} and all the \textit{Things} within a zone are connected to a single $EP^{++}$ present in \textit{Fog}.   

\textit{Fog}: The embedded processor ($EP^{++}$ -- Raspberry Pi 3 model B) present in the Fog is responsible for collecting and processing data from all the \textit{Things} present within its zone. It receives the footfall signal over ZIGBEE network, decompresses it, extracts important features from it and finally classifies the signal for human identification. The final results are stored in a local database present within the Fog.

\textit{Cloud}: It communicates only with the $EP^{++}$ over Ethernet/Wi-Fi and hosts a single database which contains information related to human movement in different zones and sub-zones. To reduce the communication bandwidth, the Cloud periodically connects with the Fog and updates its database tables. These tables present in the Cloud and Fog can be used to design different applications for attendance, surveillance, activity monitoring~\cite{bodhi_intruder}, fall detection~\cite{fall_detection_Wang}, appliance control in smart homes~\cite{jun2015controlling} etc.

The activities performed by the processors present in the \textit{Things} and \textit{Fog} are explained below.

\begin{figure}[!b]
\centering
\includegraphics[width=\linewidth]{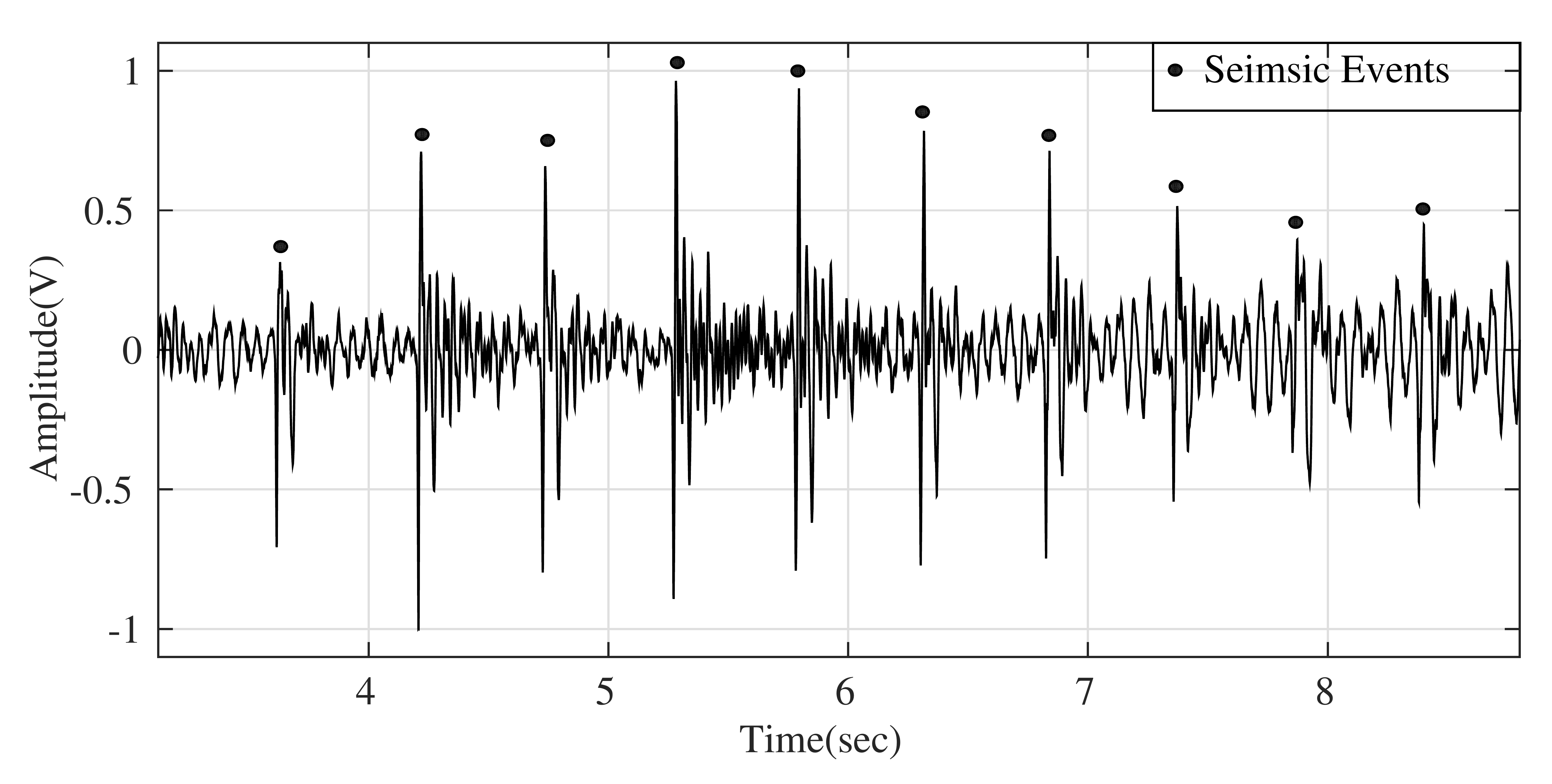}
\caption{Seismic Signal corresponding to ten consecutive footsteps. Each black dot ($\bullet$) represents a single footstep. }
\label{fig:seismic signal}
\end{figure}

\setcounter{MaxMatrixCols}{25}
\begin{figure*}[]
\begin{align*}
\mathbf{\boldsymbol{D}} & =\begin{pmatrix}
 \vdots & \vdots & \vdots & \vdots &  \vdots  & \vdots & 
 \vdots & \vdots & \vdots & \vdots &  \vdots  & \vdots & & &
 \vdots & \vdots & \vdots & \vdots &  \vdots  & \vdots 
 \\
  \rotatebox{90}{\hspace{15pt} $f(t;\tau=0)$ }  & \rotatebox{90}{$f_c(t;\tau=0,\omega=5)$} & \rotatebox{90}{$f_s(t;\tau=0,\omega=5)$} & \hdots & \rotatebox{90}{$f_c(t;\tau=0,\omega=250)$} & 
  \rotatebox{90}{$f_s(t;\tau=0,\omega=250)$} &
 \rotatebox{90}{\hspace{15pt} $f(t;\tau=\frac{1}{L-1})$}  & \rotatebox{90}{$f_c(t;\tau=\frac{1}{L-1},\omega=5)$} & \rotatebox{90}{$f_s(t;\tau=\frac{1}{L-1},\omega=5)$} & \hdots & \rotatebox{90}{$f_c(t;\tau=\frac{1}{L-1},\omega=250)$} & 
  \rotatebox{90}{$f_s(t;\tau=\frac{1}{L-1},\omega=250)$}
  & \hdots & \hdots &
    \rotatebox{90}{\hspace{15pt} $f(t;\tau=1)$ }  & \rotatebox{90}{$f_c(t;\tau=1,\omega=5)$} & \rotatebox{90}{$f_s(t;\tau=1,\omega=5)$} & \hdots & \rotatebox{90}{$f_c(t;\tau=1,\omega=250)$} & 
  \rotatebox{90}{$f_s(t;\tau=1,\omega=250)$} 
  \\
\vdots & \vdots & \vdots & \vdots & \vdots & \vdots & 
\vdots & \vdots & \vdots & \vdots & \vdots & \vdots & & &
\vdots & \vdots & \vdots & \vdots &  \vdots  & \vdots
\end{pmatrix}
\end{align*} 
\caption{The Dictionary ($\boldsymbol{D} \in \Re^{L \times (1 + 2|w|)L}$) used for compressing the footfall events by \textit{DS8BP}.}
\label{Fig:Dictionary}
\end{figure*}

\subsection{Signal Acquisition} \label{SubSec:sig_rec}

$EP$s convert the analog seismic waves into digital by using an ADC of resolution 16 bits and sampling frequency($f_s$) of 8 kHz. Fig.~\ref{fig:seismic signal} shows a seismic signal containing human footsteps. The geophone used in the paper has an indoor range of 2.5\,m and an outdoor range of 25\,m.    

\subsection{Event Extraction} 
\label{SubSec:event_extraction}

An Event extraction technique is used to detect and fetch portions of seismic signal that represent footfalls. A fixed threshold based event detection technique fails to capture the events with time varying amplitudes (refer Fig.~\ref{fig:seismic signal}). So, we used an adaptive threshold based event extraction technique for footfall signal~\cite{bodhi_intruder, Anch1701:Predicting}. It was observed that if two consecutive footfalls occur within a short period of time, the event extraction technique detects them as a single event. This happens when an individual is walking very close ($<0.25m$) to the seismic sensor and a new footfall starts before the previous one damps out. The event detection technique also picks up floor/ground vibrations of high magnitude (movements of tables and chairs) other than footfall. However, these events have a short lifespan and die out quickly. Therefore, if the width of the detected event is too large then it is actually a combination of two footfalls and if it is too small then it is actually a noise. To get rid of these types of false events, an upper bound of 0.437\,sec and a lower bound of 0.144\,sec were set on the width of the events. These bounds were set by studying different types of footfall signals.

\begin{algorithm}[!b]
\caption{Compression Algorithm: \textit{DS8BP}}
\label{Algo:compression}
\begin{algorithmic}[1]
\State $\it{Sig}$  \Comment{Input: Seismic signal (corresponding a footstep) after event extraction }
\State $L_{GC}$   \Comment{Input:  Lower threshold on the number of atoms}
\State $H_{GC}$    \Comment{Input: Higher threshold on the number of atoms}
\State $\it{Sig_{comp}}=[\ ]$   \Comment{Output: Compressed Signal }
\State $\it{I}=[ \ ]$   \Comment{Output: Atom Index }
\State $\it{L}=[ \ ]$   \Comment{Output: Length of the original signal}

\State $L = length(Sig)$
\State  $\textbf{D}= \textit{generate\_dictionary}(L)$
\State  $Sig_{ds}= \textit{Downsample}(Sig,8)$
\State $\hat x = \underset{x}{min}  ||\textbf{D}x - Sig_{ds}||_2 + ||x||_1 $
\State  $ [ \hat x_{sort}, \  \hat x_{index} ]= sort(\hat x,decreasing)$
\State $\hat x_{{energy}} = \sum_{i}(\hat x_{sort_i})^2$
\State $\hat x_{sortL }= \hat x_{sort }(1:L-1) $

\State $\hat x_{sortE} = \frac{Lower\_Tri(L-1,L-1)*\hat x_{sortL }}{\hat x_{{energy}}} $
\State $ \hat i= \max_i \  s.t. \ \sum_{i} \hat x_{sortE} < 1$
\State $I = x_{index}(1: \hat i)$

\If {$ L_{GC} \leq length(I) \leq H_{GC}$}
	\State $\widehat{\textbf{D}} = \textbf{D}(:,I)$
    \State $Sig_{comp}=\widehat{\textbf{D}}^{\dagger}*Sig_{ds}$
\Else
  \State Discard $Sig$
\EndIf

\end{algorithmic}
\end{algorithm}

\subsection{Signal Compression and Decompression} \label{SubSec:sig_compress_decompress}



$EP$'s sample the seismic signals at 8kHz  i.e each of them generates 8000 sample points per second. Transmitting data at such a high rate will require a high bandwidth and consume a lot of power. So, we proposed a basis pursuit~\cite{comp_sen_WSN} based compression technique (\textit{DS8BP}) to reduce the length of the signal. Down-sampling the raw signal is also an alternative to decrease signal length, however the proposed technique has capability to compress the signal much below the down-sampled signal in spite of maintaining an acceptable recovery error (refer Section~\ref{SubSec:pref_DS8BP}).


\textit{DS8BP} projects a down sampled version ($Sig_{ds}$) of a finite length seismic event ($Sig$) on an over complete dictionary $\boldsymbol{D}$. $\boldsymbol{D}$ is selected in such a manner that footfall events can be expressed by linear combinations of few columns (atoms) of $\boldsymbol{D}$ i.e coefficients of the columns of $D$ that represent $Sig_{ds}$ are always sparse in nature. It was observed that footfall signals can be represented by linear combinations of Gabor functions. Fig.~\ref{Fig:Dictionary} shows the dictionary $\boldsymbol{D} (\in \Re^{L \times (1 + 2|w|)L})$,  where $f(t,\tau) = e^{ \frac{(t-\tau)^2}{\sigma^2}}$, $f_c(t,\tau,\omega) = e^{ \frac{(t-\tau)^2}{\sigma^2}}cos(\omega t)$, $f_s(t,\tau,\omega) = e^{ \frac{(t-\tau)^2}{\sigma^2}}sin(\omega t)$, $\omega = \{ 5,10,15, \hdots, 250 \}$, $\sigma = 0.5 \text{ (set heuristacally)}$, $t=\tau = \{ 0,\frac{1}{L-1},\hdots, 1 \}$, $|t|=|\tau| = L$ and $L$= length of the down-sampled signal $Sig_{ds}$ ($|.|$ represents the cardinality of the set). To generate the dictionary we only need to specify the value of $L$. Maximum value of $\omega$ is 250 Hz, as it is sufficient for representation of a footfall event.

Algorithm\ref{Algo:compression} shows the pseudocode for implementing \textit{DS8BP}. $\hat x$ represents coefficients of the columns in D required for decompressing the signal($Sig_{ds}$). $\hat x$ is obtained using LASSO regularization (Line 10 in Algorithm~\ref{Algo:compression}), so it is always sparse in nature. A new matrix $\widehat D$ is formed by eliminating the columns of $D$ whose corresponding coefficients in $\hat x$ are negligible. An energy based technique (Line 9 \-- 14 in Algorithm~\ref{Algo:compression}) is used to find the columns (Atom index($I$) ) of D. Seismic events ($Sig$) are discarded if the number of atoms required to represent the signal is lower than $L_{GC}$ or higher than $H_{GC}$ (refer Section.~\ref{SubSec:pref_DS8BP}). Finally, the compressed signal ($Sig_{comp}$) is obtained by projecting the down-sampled signal ($Sig_{ds}$) on the matrix $\widehat D$. The original signal is down-sampled before performing the compression technique to reduce the size of the dictionary. LASSO regularisation and pseudoinverse ($\dagger$) cannot be performed within an $EP$ if the size of $D$ (proportional to $L$ -- refer Fig.~\ref{Fig:Dictionary}) is very large. The compressed footfall event $Sig_{comp}$ is transmitted to $EP^{++}$ using XBee. 
$EP^{++}$ recovers the original signal using Algorithm~\ref{Algo:de_compression}. For decompression we require following three quantities \--- a) compressed signal ($Sig_{comp}$), b) position of the significant atoms ($I$) and, c) length of the down-sampled signal ($L$)  (to generate dictionary $\textbf{D}$). Using this information the original signal can be recovered by weighted summation of the columns of $\textbf{D}$ indexed by $I$. The elements in vector $Sig_{comp}$ represent the weight by which the selected columns of $D$ are multiplied.
\begin{algorithm}[!h]
\caption{De-Compression Algorithm}
\label{Algo:de_compression}
\begin{algorithmic}[1]
\State $\it{Sig_{comp}}$   \Comment{Input: Compressed Signal }
\State $\it{I}$   \Comment{Input: Atom Index }
\State $\it{L}$   \Comment{Input: Original Signal Length }
\State $\it{Sig_{rec}=[ \ ]}$  \Comment{Output: Recovered Signal}
\State  $\textbf{D}= \textit{generate\_dictionary}(L)$
\State  $\widehat {\textbf{D}} = \textbf{D}(:,I)$
\State $\it{Sig_{rec}}= \widehat {\textbf{D}} * Sig_{comp}$
\end{algorithmic}
\end{algorithm}

If $M$ is the length of the compressed signal ($Sig_{comp}$) then $10M +2 $ bytes ($M \times 8 + M \times 2 + 2$) of data are required to transmit a single footfall event. Elements of $Sig_{comp}$ are represented by 4 bytes (double precision floating point format) whereas $I$ and $L$ are represented by 2 bytes.

%

 
%

\subsection{Feature Engineering} \label{SubSec:feature_engineering}

Features are the most vital factor in any biometric system. We collected both time and frequency domain related features from the decompressed footfall signals. The length of a footfall and the gap between two consecutive footsteps (cadence) were also included in the feature vector. Features related to the energy of the signal in time domain were not considered, in order to make the feature set distance (between the person and the sensor) independent. From Fig.~\ref{fig:seismic signal}, it can be seen that the amplitudes of consecutive footfalls do not remain the same. Therefore, a person walking close or far from the geophone will make no difference until they are within the sensing range of the sensor. The extracted features are shown in Table~\ref{tab:features_table}. 
\begin{table}[!t]
\centering
\scriptsize
\caption{Features extracted from a single seismic event.}
\begin{tabular}{|c|c|c|}
\hline
\textbf{Features}                           & \textbf{Parameter}                      & \textbf{No. of features} \\ \hline
\multirow{3}{*}{\textbf{Time Domain}}       & Standard Deviation                      & 1                        \\ \cline{2-3} 
                                            & Skewness                                & 1                        \\ \cline{2-3} 
                                            & Kurtosis                                & 1                        \\ \hline
\multirow{4}{*}{\textbf{Hilbert Transform}} & Mean                                    & 1                        \\ \cline{2-3} 
                                            & Standard Deviation                      & 1                        \\ \cline{2-3} 
                                            & Skewness                                & 1                        \\ \cline{2-3} 
                                            & Kurtosis                                & 1                        \\ \hline
\multirow{4}{*}{\textbf{Frequency}}         & Energy in 0-2 Hz.                       & 1                        \\ \cline{2-3} 
                                            & Energy in 2-4 Hz.                       & 1                        \\ \cline{2-3} 
                                            & $\vdots$                                   & $\vdots$                       \\ \cline{2-3} 
                                            & Energy in 248-250 Hz.                   & 1                        \\ \hline
\textbf{Gait Property 1}                    & Cadence (rhythm of footsteps)           & 1                        \\ \hline
\textbf{Gait Property 2}                    & Duration of a footstep(in samples) & 1                        \\ \hline
\multicolumn{2}{|c|}{\textbf{Total Features}}                                         & \textbf{135}             \\ \hline
\end{tabular}

\label{tab:features_table}
\end{table}

\subsection{Classification and Data storage} \label{SubSec:classification}
 
Trained models are preloaded in $EP^{++}$ to predict the identity of individuals using the feature vectors. The final results are stored in a local database within $EP^{++}$. $EP^{++}$ uploads the results to the central database in the cloud\cite{think_speak} periodically.

\begin{table}[!b]
\caption{Details of the Data Set.}
\centering
\begin{tabular}{|c|c||c|c|}
\hline
\begin{tabular}[c]{@{}c@{}}Class / \\ Person\end{tabular} &  No. of Footstep & \begin{tabular}[c]{@{}c@{}}Class /\\ Person\end{tabular}  & No. of Footstep \\ \hline
1 & 5994 & 5 &  5431 \\ \hline
2 & 6251 & 6 &  5554\\ \hline
3 & 5299 & 7 &  6091\\ \hline
4 & 5809 & 8 &  6060\\ \hline
\end{tabular}
\label{Tb:dataset}
\end{table}
\section{Results and Discussion} \label{Sec:result_diss}
Footstep data of eight individuals (four males and four females)  were collected using a geophone of 2.88\,V/mm/sec sensitivity and 100 gain. Average age of the individuals was between 20 to 25 years. The sensor was placed on the floor of the lab and each individual was made to walk (barefooted) in circles of radius 1\,m to 2.5\,m around it. The dataset consisted more than one hour of recorded seismic signals of each individual (collected over a period of one month). Footfall events were extracted from the signal using the adaptive threshold based event extraction technique (Section~\ref{SubSec:event_extraction}). The final dataset was created by manually annotating the extracted footfall events. The details of the dataset are shown in Table\,\ref{Tb:dataset}. To the best of our knowledge, this is the largest footstep database collected to date containing 46,489 footsteps. Certain criteria required for hardware implementation were studied: best suited classifier, optimal sampling frequency, and number of training data (footfalls) required per individual. Also it is difficult to achieve high prediction accuracy by using features only from a single footstep, as the feature space of different individuals remains overlapped. So, we have considered the mean of the features extracted from $F$ consecutive footsteps as a single sample. A study has also been conducted to obtain the optimal value of $F$. Performances\footnote{All  the computations  were  performed  using a  64  bit  operating  system running on an Intel(R) Core(TM) i7-4790 (CPU @ 3.60GHz x 8 ) processor with 16.00  GB  RAM} of different multiclass classifiers~\cite{Pattern_recognition_bishop} (Logistic Regression (LR), SVM linear (SVM-Lin), SVM Gaussian (SVM-RBF), single hidden layer Artificial Neural Network (ANN), Linear Discriminate Analysis (LDA)) were tested using the dataset (Table\,\ref{Tb:dataset}).

\subsection{Selection of Optimal Number of footsteps/sample  } \label{SubSec:optimal_no_footstep}


Table~\ref{Tb:accu_foot_per_sample} shows the performance of different classifiers on datasets of varying footsteps per sample. It can be observed that the performance of the classifiers escalates as the number of footstep/sample increases. SVM-RBF outperformed the rest of the classifiers when single footstep/sample was considered by achieving an accuracy of 71.2\%. Its performance was $11\%$ better than LR and $7\%$ better than SVM-Lin. It proved that the classes were not linearly separable in the feature space of 1 footstep/sample. There was a drastic increment in performance ( 9\% for SVM RBF and 11\%-12\% for the rest of the classifiers) as we moved from 1 to 2 footstep per sample, as in case of 1 footstep/sample the feature cadence was absent. However, there was only 5\%-6\% improvement in accuracy in 3 footsteps/sample over 2 footsteps/sample. We achieved an accuracy of 90\,\% with SVM-RBF by considering 5 footsteps/sample compared to 83\,\% achieved in~\cite{Pan_Indoor_Person_Identification}. They also considered five footsteps of highest SNR as a sample. As more number of footsteps were considered in a single sample, classes in the feature space become more distinct. All the classifiers had similar accuracies (about $97\%$) when 25 footsteps/sample were considered. In that case, the feature space of the classes became linearly separable and all machine learning techniques (even those not used in the paper) could easily classify the data with high accuracy. 

The selection of an optimal number of footstep depends on the type of application for which the system is being used. However, 7 footsteps/sample is a good choice in terms of accuracy and data acquisition, as it is possible to collect 7 consecutive footsteps using a single geophone (sensing diameter of a geophone in indoor is 5~m approximately). 15-20 footsteps per sample can also be be used in applications which require very high accuracy. For collecting 15-20 consecutive footsteps in indoor condition an array of seismic sensors can to be used. 

\begin{table}[]
\centering
\caption{Accuracy of different classifiers obtained by varying the number of footsteps/sample. ($f_s = 8 kHz$)}
\label{Tb:accu_foot_per_sample}
\resizebox{.49\textwidth}{!}{\begin{tabular}{|c|c|c|c|c|c|c|c|c|}
\hline
\multirow{2}{*}{\textbf{\begin{tabular}[c]{@{}c@{}}Classifiers\\ Accuracy(\%)\end{tabular}}} & \multicolumn{8}{c|}{\textbf{Number of Footsteps/sample}} \\ \cline{2-9} 
 & \textbf{1} & \textbf{2} &\textbf{3} & \textbf{5} & \textbf{7} & \textbf{10} & \textbf{15} & \textbf{25} \\ \hline
\textbf{SVM Lin.} & 64.5 & 76.4 & 82.0 & 88.2 & 91.1 & 94.4 & 96.6 & 98.0  \\ \hline
\textbf{SVM RBF} & 71.2 & 80.0 & 85.3 & 90.4 & 92.2 & 95.3 & 96.1 & 98.1 \\ \hline
\textbf{LR} & 60.9 & 73.9 & 79.4 & 85.7 & 90.5 & 93.5 & 96.0 & 97.3 \\ \hline
\textbf{LDA} & 54.5 & 65.8 & 71.8 & 80.1 & 84.8 & 88.4 & 93.8 & 96.2\\ \hline
\textbf{ANN} & 61.9 & 73.8 & 80.3 & 85.9 & 89.6 & 92.3 & 95.1 & 97.4 \\ \hline
\end{tabular}}
\end{table}

\subsection{Selection of best suitable Classifiers for person identification  } \label{Sec:best_classifier}

The performances of the classifiers were obtained using a dataset where each sample represented features from 7 consecutive footsteps ($F=7$) sampled at 8~kHz. 10 fold cross validation technique was used to avoid over and underfitting of the classifiers. The test set was normalized by the mean and standard deviation obtained by normalizing the training set to imitate real time scenarios. The performance parameters used for classifier selection were accuracy, precision, recall, F1 score, and computational complexity.

It was observed that SVM-RBF outperformed rest of the techniques by achieving a prediction accuracy of 92.29\%. SVM-RBF was followed by SVM-Lin and LR, by achieving an accuracy of 91.90\% and 90.13\%. Out of all the classifier, LDA's performance was the worst (accuracy=84.76\%). It is due to the reason that LDA assumes the features are drawn from a Gaussian distribution and all the classes share a common covariance matrix. Out of all the classifiers, LR was computationally least expensive as it separated the classes in the dataset with straight lines. ANN performed better than LDA (accuracy of 88.97\%), however it required a lot of computational power for model training.

\subsection{Selection of Optimal Sampling Frequency} \label{Sec:samp_freq_opt}
 
\begin{figure}[]
\centering
\includegraphics[width=\linewidth]{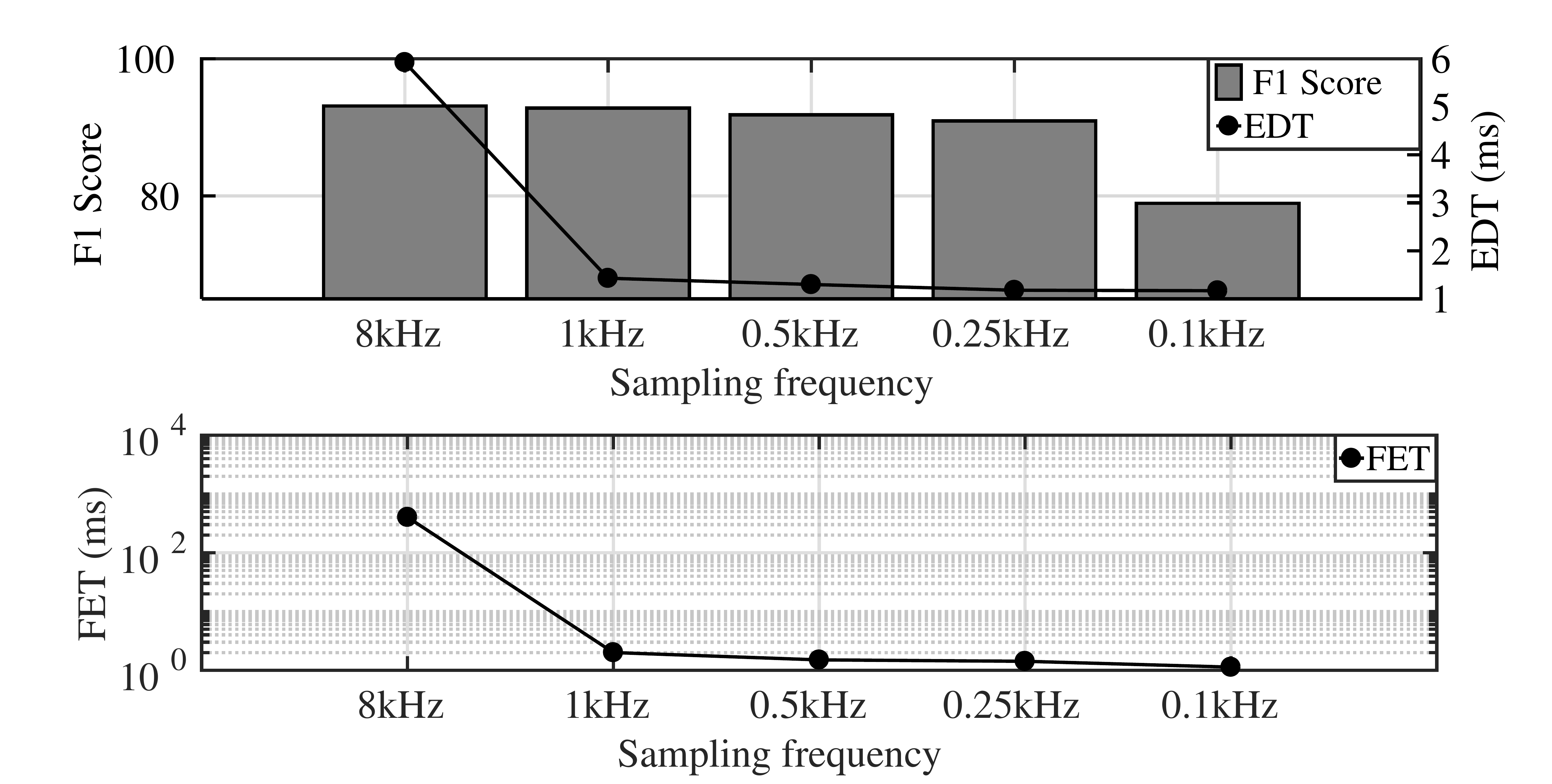}
\caption{Variation of F1 score, Event Detection time (EDT) and Feature extraction time (FET) of the signal with change in $f_s$. (Classifier used SVM-RBF \& each sample contains 7 footsteps)}
\label{fig:accu_fs}
\end{figure}

The seismic waveform generated by footfalls are low frequency signals and can be sampled at a much lower sampling rate($f_s$) than 8 kHz. Reduction in the sampling rate decreases event detection time (EDT: total time to detect and extract all the events of a finite length signal divided by the total number of events) and feature extraction time per sample (FET). However, reducing the sampling frequency below a certain point deteriorates classifiers accuracy as downsampling affects the integrity of the signal. So, we have tried to obtain a sampling frequency below 8 kHz where the accuracy of the classifier remains unaltered. In this study, to obtain the optimal sampling frequency, we used SVM-RBF on a dataset containing 7 consecutive footsteps per sample. It was observed (refer Fig.~\ref{fig:accu_fs}) that accuracy remained unaltered (decreased by less than 2\%) when the sampling frequency was reduced from 8 kHz to 500 HZ whereas the EDT and FET reduced by 78\% and 99\%.

\begin{figure}[!t]
\centering
\includegraphics[width=\linewidth]{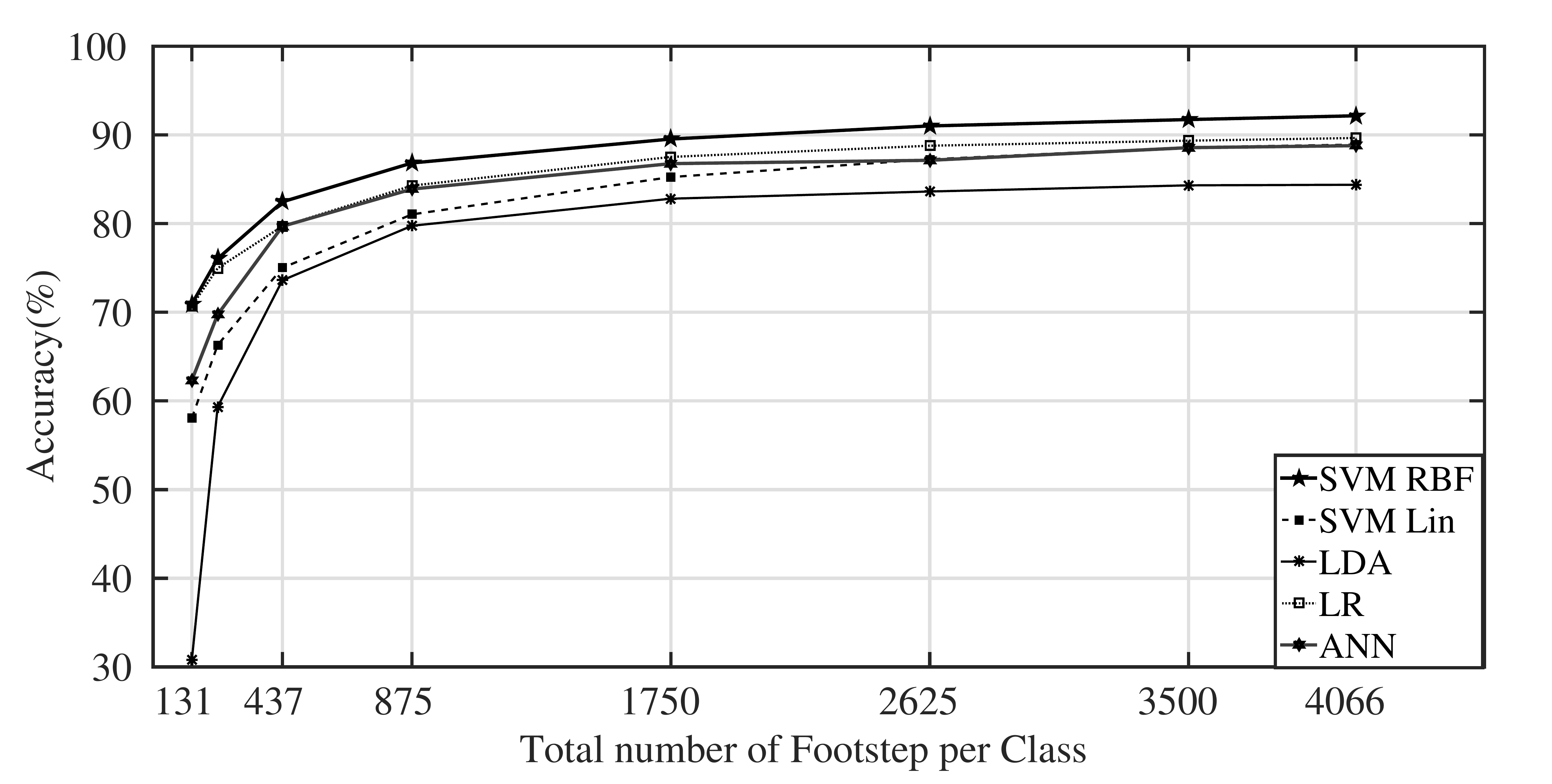}
\caption{Learning Curve of all the classifiers on seismic data set. (Each sample has 7 footsteps, $f_s$=8 kHz)}
\label{fig:learning_curve}
\end{figure}

\subsection{Selection of Optimal Number of Training samples per class} \label{Sec:train_sample_opt}

From an implementation point of view, it is important to know the required number of training samples per class to obtain the desired classification accuracy. The number of training samples can be obtained from the learning curve of the classifiers (refer Fig.~\ref{fig:learning_curve}). It can be found that to achieve an accuracy greater than 85\% we need 875 footsteps per class i.e 8 minutes of walking (assuming a normal human has approximately 100 footsteps per minute). But to get an accuracy of 90\%, 17 minutes of seismic data are required. SVM-RBF has the highest learning rate compared to the other classifiers followed by LR, whereas ANN has the poorest learning rate. Using the learning curve (Fig.~\ref{fig:learning_curve}) a designer can easily pick an appropriate value for the number of footstep required to train his/her classifier.   

In spite of the fact that LR has the least prediction time and has almost similar performance parameters to SVM-RBF, the latter is the best suited classifier for footstep data. SVM-RBF has better performance parameters (even when samples have lower number of footsteps: 71.2\% for 1 footstep/sample) and a better learning curve than the others. In case of single footstep/sample, the accuracy of LR is only 60\%. Table~\ref{Tb:conf_svm-rbf} shows the confusion matrix of a test instance of SVM-RBF classifier with 7 footsteps/sample and 500Hz as the sampling frequency. 
%
\begin{table}[!h]
\centering
\caption{Confusion Matrix of the SVM-RBF for 7 footsteps/sample and 500 Hz sampling frequency.}
    \noindent\begin{tabular}{cc*{8}{|E}|}
\cline{3-10}
\multirow{8}{*}{\rotatebox[origin=c]{90}{\textbf{Predicted Class}}}
&\textbf{P1} & 0.94 & 0 & 0 & 0 & 0 & 0.01 & 0 & 0 \\ \cline{3-10}
&\textbf{P2} & 0 & 0.99 & 0 & 0 & 0 & 0 & 0 & 0  \\ \cline{3-10}
&\textbf{P3} & 0.01 & 0 & 0.98 & 0.06 & 0 & 0.01 & 0 & 0.05  \\ \cline{3-10}
&\textbf{P4} & 0.04 & 0 & 0 & 0.98 & 0 & 0 & 0 & 0  \\ \cline{3-10}
&\textbf{P5} & 0 & 0 & 0 & 0 & 0.94 & 0.03 & 0.02 & 0 \\ \cline{3-10}
&\textbf{P6} & 0 & 0 & 0 & 0.02 & 0.05 & 0.93 & 0 & 0.03  \\ \cline{3-10}
&\textbf{P7} & 0 & 0 & 0 & 0 & 0 & 0 & 0.95 & 0  \\ \cline{3-10}
&\textbf{P8} & 0 & 0.01 & 0.02 & 0 & 0.02 & 0.02 & 0.03 & 0.92 \\ \cline{3-10}

& \multicolumn{1}{c}{} & \multicolumn{1}{c}{\textbf{P1}} & \multicolumn{1}{c}{\textbf{P2}} 
  & \multicolumn{1}{c}{\textbf{P3}} & \multicolumn{1}{c}{\textbf{P4}} & \multicolumn{1}{c}{\textbf{P5}} 
  & \multicolumn{1}{c}{\textbf{P6}} & \multicolumn{1}{c}{\textbf{P7}} & \multicolumn{1}{c}{\textbf{P8}}\\ 
 & \multicolumn{1}{c}{} &    \multicolumn{8}{c}{\textbf{Actual Class}} 
\end{tabular}
\label{Tb:conf_svm-rbf}
\end{table}

\subsection{Performance of the proposed Compression and De-compression Technique} \label{SubSec:pref_DS8BP}

The proposed technique \textit{DS8BP} (discussed in Section.~\ref{SubSec:sig_compress_decompress}) used for signal compression first down samples the 8 kHz ($Sig$) signal to 1 kHz ($Sig_{ds}$)) and then uses basis pursuit technique to decompress it. The signals which have low compression factors (length $Sig_{comp}$/length $Sig_{ds}$) need large number of basis functions (i.e columns of $D$) for representation. Footsteps are low frequency signals and can be represented by a handful of basis functions. A seismic event with very low compression factor is not a footfall, they might be high amplitude noise wrongly picked up by the event extraction technique. Noise constitutes different frequencies from low to high and requires a large number of basis function for representation. Hence, signals whose compression factors were less than 5 ($L_{GC}$) were eliminated. Also, those signals that had very high compression factor above 40 ($H_{GC}$) were not recoverable as they had very few basis functions. \textit{DS8BP} acts as a filter for eliminating signals falsely picked up by the event extraction technique. The values of $L_{GC}$ and $H_{GC}$ were set experimentally.   

The compression technique \textit{DS8BP} was compared with another technique named \textit{DS16}. \textit{DS16} decimated the seismic signal by a factor of 16. The decimation factor was kept at 16 (i.e. sampling frequency 500 Hz) as it was the minimum sampling frequency upto which the integrity of the signal remained intact (refer Sec.\ref{Sec:samp_freq_opt}). Average length of a footfall event was around 250.62 millisecond (i.e  2005 samples/event when sampled at 8 kHz). By using the footfall events in the dataset (Table ~\ref{Tb:dataset}), it was observed that in case of \textit{DS8BP}, the average compression factor (length $Sig_{comp}$/length $ Sig_{ds}$) of a single footstep was 13.54 and the average length of a compressed footfall event ($Sig_{comp}$) was 18.51 ($M$). It was also noticed that \textit{DS8BP} compresses the original 8 kHz signal by a factor of 108.32 ($8 \times 13.54$) whereas \textit{DS16} compresses only by a factor of 16. 
A wireless transmitter (XBee 868LR) with a data rate of 80 kbps will requires 18.71 millisecond \ $(\frac{(10M + 2) \times 8}{80 \times 1000})$ and 100 millisecond $(\frac{(125 \times 8) \times 8}{80 \times 1000})$ to transmit a single footfall event when compressed by \textit{DS8BP} and \textit{DS16}.   
\begin{figure}[!t]
    \centering
    \includegraphics[width=\linewidth]{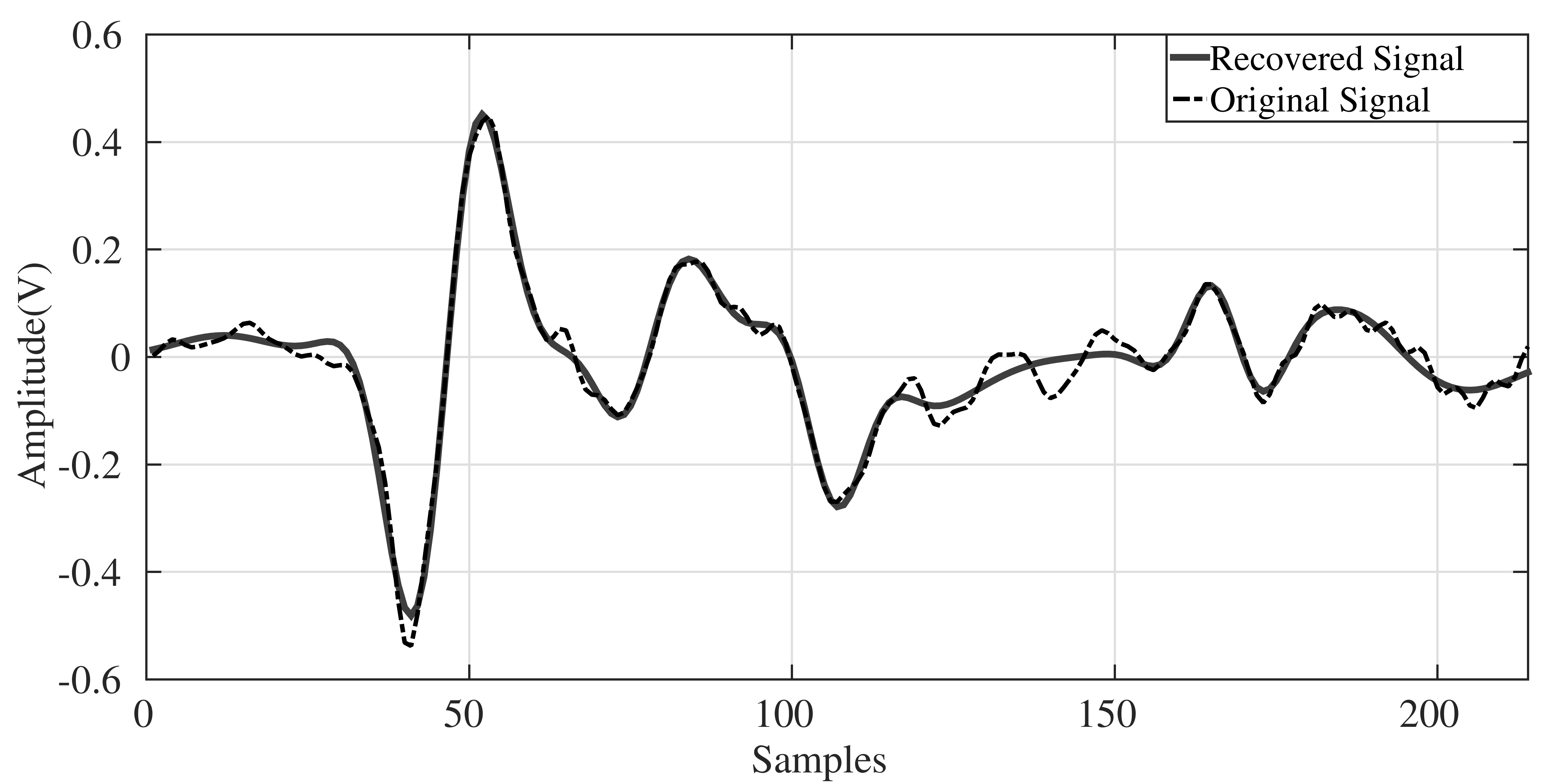}
    \caption{ The original ($Sig$) and the decompressed signal ($Sig_{rec}$) of a footfall event. The compression and decompression were performed using Algorithm~\ref{Algo:compression} and Algorithm~\ref{Algo:de_compression}.}
    \label{fig:orig_recov}
\end{figure}
\begin{figure}[!t]
\centering
\includegraphics[width=\linewidth]{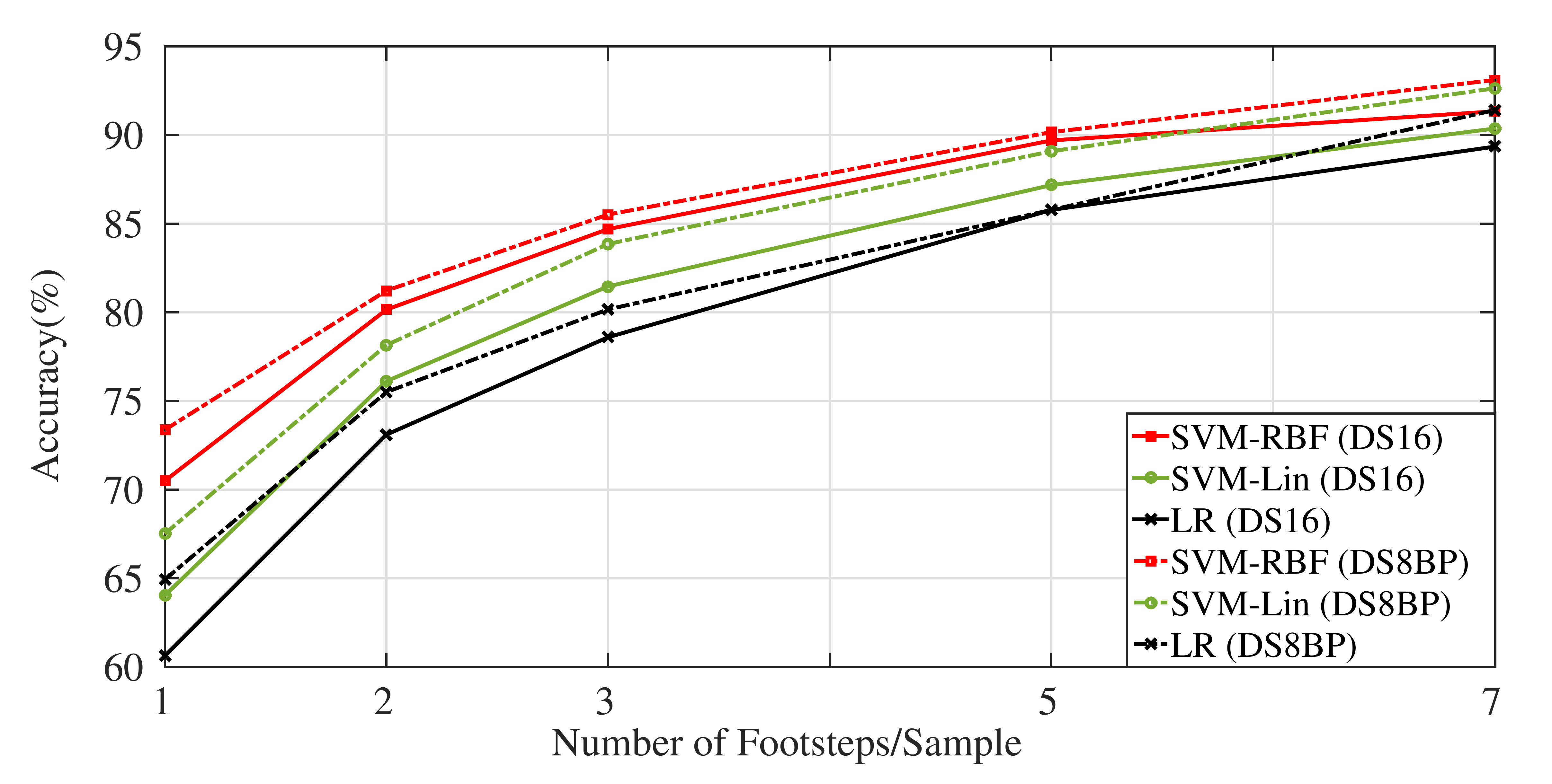}
\caption{Performance analysis of classifiers on recovered signals after being compressed by $DS16$ and $DS8BP$.} 
\label{fig:cs_noncs_acc}
\end{figure}

 Fig.~\ref{fig:orig_recov} shows the original ($Sig_{ds}$), and recovered footfall event($Sig_{rec}$) corresponding to a single footstep obtained using \textit{DS8BP}. It can be seen that the recovered signal completely matches with the original signal and does not contain high frequency components. The recovered signal is formed by a linear combination of a handful of Gabor functions (20 in this case of Fig.~\ref{fig:orig_recov}), with a maximum frequency of 250 Hz. So there is no basis that represents the high frequency components of the original signal. Fig.~\ref{fig:cs_noncs_acc} shows the effect of the compression techniques on classifiers' performance. The solid lines and the dotted lines represent the accuracies of classifiers (SVM-RBF, SVM-Lin, LR) on decompressed signals. The signals are compressed by using \textit{DS16} and \textit{DS8BP} respectively. Fig.~\ref{fig:cs_noncs_acc} shows the accuracies of the classifiers w.r.t the number of footsteps per sample. In all the three classifiers, signals compressed by \textit{DS8BP} technique achieved higher classification accuracy as compared to signals compressed by \textit{DS16}. This is due to the fact that \textit{DS8BP} de-noises the signals as it works as a smoothing function. The accuracy of SVM-RBF, SVM-Lin, and LR increases by (2\%-4.3\%) in case of 1 footstep/sample and (.8\%-2.4\%) with 3 footsteps/sample when \textit{DS8BP} is used instead of \textit{DS16}. 
%

\section{Conclusion and Future Work } \label{Sec:future_con}

In this paper, we have implemented a Fog computing architecture based person identification system using seismic events generated from their footfalls. The following observations are made \--- clubbing consecutive footsteps per sample increases the prediction accuracy, downsampling the 8 kHz signal to 500 Hz does not affect classifiers' accuracy,  although it reduces FET(feature extraction time) and EDT(event detection time). The proposed signal compression technique \textit{DS8BP} reduces the 8 kHz seismic signal by 108 times and increases the classifiers' accuracy by (2-4)\% in 1 footstep/sample scenario. \textit{DS8BP} also increases the security of the system as its impossible to decode the compressed signal without knowing the dictionary ($D$).  

The proposed set up for human identification assumes that a single person was present within the active region of the sensor. This assumption is valid in certain indoor scenario, however in an outdoor environment there are chances that more than one person can step into the active arena of the geophone. So, techniques that can separate combined footstep signals of two (or more) individuals might be explored in future. As this work has achieved very high accuracy for person identification, it can be extended to detecting intruder i.e. non registered persons (anomaly detection). Its applications can be further extended to surveillance, border monitoring etc. Deep Neural algorithms: Convolution Neural Network, Recurrent Neural Network (RNN) can be used for feature extraction and classification.

 \bibliographystyle{IEEEtran}
 \bibliography{reference}

\begin{thebibliography}{10}
\providecommand{\url}[1]{#1}
\csname url@samestyle\endcsname
\providecommand{\newblock}{\relax}
\providecommand{\bibinfo}[2]{#2}
\providecommand{\BIBentrySTDinterwordspacing}{\spaceskip=0pt\relax}
\providecommand{\BIBentryALTinterwordstretchfactor}{4}
\providecommand{\BIBentryALTinterwordspacing}{\spaceskip=\fontdimen2\font plus
\BIBentryALTinterwordstretchfactor\fontdimen3\font minus
  \fontdimen4\font\relax}
\providecommand{\BIBforeignlanguage}[2]{{%
\expandafter\ifx\csname l@#1\endcsname\relax
\typeout{** WARNING: IEEEtran.bst: No hyphenation pattern has been}%
\typeout{** loaded for the language `#1'. Using the pattern for}%
\typeout{** the default language instead.}%
\else
\language=\csname l@#1\endcsname
\fi
#2}}
\providecommand{\BIBdecl}{\relax}
\BIBdecl

\bibitem{fang2006path}
J.~S. Fang, Q.~Hao, D.~J. Brady, M.~Shankar, B.~D. Guenther, N.~P. Pitsianis,
  and K.~Y. Hsu, ``Path-dependent human identification using a pyroelectric
  infrared sensor and fresnel lens arrays,'' \emph{Optics express}, vol.~14,
  no.~2, pp. 609--624, 2006.

\bibitem{wang2003silhouette}
L.~Wang, T.~Tan, H.~Ning, and W.~Hu, ``Silhouette analysis-based gait
  recognition for human identification,'' \emph{IEEE Trans. Pattern Anal. Mach.
  Intell.}, vol.~25, no.~12, pp. 1505--1518, 2003.

\bibitem{jain2000biometric}
A.~Jain, L.~Hong, and S.~Pankanti, ``Biometric identification,''
  \emph{Communications of the ACM}, vol.~43, no.~2, pp. 90--98, 2000.

\bibitem{Begg_svm_gait}
R.~K. Begg, M.~Palaniswami, and B.~Owen, ``Support vector machines for
  automated gait classification,'' \emph{IEEE Trans. Biomed. Eng.}, vol.~52,
  no.~5, pp. 828--838, May 2005.

\bibitem{iris_lie}
N.~Liu, J.~Liu, Z.~Sun, and T.~Tan, ``A code-level approach to heterogeneous
  iris recognition,'' \emph{IEEE Trans. Inf. Forensics Security}, vol.~12,
  no.~10, pp. 2373--2386, Oct 2017.

\bibitem{Wang}
X.~Wang, T.~Yang, Y.~Yu, R.~Zhang, and F.~Guo, ``Footstep-identification system
  based on walking interval,'' \emph{IEEE Intelligent Systems}, vol.~30, no.~2,
  pp. 46--52, Mar 2015.

\bibitem{body_shape_mm_wave_person}
E.~Gonzalez-Sosa, R.~Vera-Rodriguez, J.~Fierrez, and V.~M. Patel, ``Exploring
  body shape from mmw images for person recognition,'' \emph{IEEE Trans. Inf.
  Forensics Security}, vol.~12, no.~9, pp. 2078--2089, Sept 2017.

\bibitem{carpet_Zhou}
B.~Zhou, M.~S. Singh, S.~Doda, M.~Yildirim, J.~Cheng, and P.~Lukowicz, ``The
  carpet knows: Identifying people in a smart environment from a single step,''
  in \emph{IEEE Int. Conf. Pervasive Computing and Communications Workshops
  (PerCom Workshops)}, March 2017, pp. 527--532.

\bibitem{jun2015controlling}
\BIBentryALTinterwordspacing
H.~JUN, ``Controlling electronic devices based on footstep pattern,'' Mar.~12
  2015, uS Patent App. 14/481,878. [Online]. Available:
  \url{https://www.google.com.pg/patents/US20150073568}
\BIBentrySTDinterwordspacing

\bibitem{Preden_fog_mist}
J.~S. Preden, K.~Tammemäe, A.~Jantsch, M.~Leier, A.~Riid, and E.~Calis, ``The
  benefits of self-awareness and attention in fog and mist computing,''
  \emph{Computer}, vol.~48, no.~7, pp. 37--45, July 2015.

\bibitem{fog_computing_Tang}
B.~Tang, Z.~Chen, G.~Hefferman, S.~Pei, T.~Wei, H.~He, and Q.~Yang,
  ``Incorporating intelligence in fog computing for big data analysis in smart
  cities,'' \emph{IEEE Trans. Ind. Informat.}, vol.~13, no.~5, pp. 2140--2150,
  Oct 2017.

\bibitem{Addlesee}
M.~D. Addlesee, A.~Jones, F.~Livesey, and F.~Samaria, ``The orl active floor
  [sensor system],'' \emph{IEEE Pers. Commun.}, vol.~4, no.~5, pp. 35--41, Oct
  1997.

\bibitem{middleton_floor}
L.~Middleton, A.~A. Buss, A.~Bazin, and M.~S. Nixon, ``A floor sensor system
  for gait recognition,'' in \emph{Fourth IEEE Workshop Automatic
  Identification Advanced Technologies (AutoID'05)}, Oct 2005, pp. 171--176.

\bibitem{SUUTALA200821}
J.~Suutala and J.~Röning, ``Methods for person identification on a
  pressure-sensitive floor: Experiments with multiple classifiers and reject
  option,'' \emph{Information Fusion}, vol.~9, no.~1, pp. 21 -- 40, 2008,
  special Issue on Applications of Ensemble Methods.

\bibitem{Qian}
G.~Qian, J.~Zhang, and A.~Kidane, ``People identification using floor pressure
  sensing and analysis,'' \emph{IEEE Sensors J.}, vol.~10, no.~9, pp.
  1447--1460, Sept 2010.

\bibitem{Miyoshi_microphone}
M.~Miyoshi, K.~Mori, Y.~Kashihara, M.~Nakao, S.~Tsuge, and M.~Fukumi,
  ``Personal identification method using footsteps,'' in \emph{SICE Annual
  Conf.}, Sept 2011, pp. 1615--1620.

\bibitem{Yun}
J.~Yun, ``User identification using gait patterns on ubifloorii,''
  \emph{Sensors}, vol.~11, no.~3, pp. 2611--2639, 2011.

\bibitem{Rodriguez_footstep_mat}
R.~Vera-Rodriguez, J.~S.~D. Mason, J.~Fierrez, and J.~Ortega-Garcia,
  ``Comparative analysis and fusion of spatiotemporal information for footstep
  recognition,'' \emph{IEEE Trans. Pattern Anal. Mach. Intell.}, vol.~35,
  no.~4, pp. 823--834, April 2013.

\bibitem{Pan_Indoor_Person_Identification}
S.~Pan, N.~Wang, Y.~Qian, I.~Velibeyoglu, H.~Y. Noh, and P.~Zhang, ``Indoor
  person identification through footstep induced structural vibration,'' in
  \emph{Int. Workshop Mobile Computing Systems Applications}, ser.
  HotMobile.\hskip 1em plus 0.5em minus 0.4em\relax ACM, 2015, pp. 81--86.

\bibitem{serra}
R.~Serra, D.~Knittel, P.~D. Croce, and R.~Peres, ``Activity recognition with
  smart polymer floor sensor: Application to human footstep recognition,''
  \emph{IEEE Sensors J.}, vol.~16, no.~14, pp. 5757--5775, July 2016.

\bibitem{Pan_2017}
\BIBentryALTinterwordspacing
S.~Pan, T.~Yu, M.~Mirshekari, J.~Fagert, A.~Bonde, O.~J. Mengshoel, H.~Y. Noh,
  and P.~Zhang, ``Footprintid: Indoor pedestrian identification through ambient
  structural vibration sensing,'' \emph{Proc. ACM Interact. Mob. Wearable
  Ubiquitous Technol.}, vol.~1, no.~3, pp. 89:1--89:31, Sep. 2017. [Online].
  Available: \url{http://doi.acm.org/10.1145/3130954}
\BIBentrySTDinterwordspacing

\bibitem{jain_biometric}
A.~K. Jain, A.~Ross, and S.~Prabhakar, ``An introduction to biometric
  recognition,'' \emph{IEEE Trans. Circuits Syst. Video Technol.}, vol.~14,
  no.~1, pp. 4--20, Jan 2004.

\bibitem{fog_face}
P.~Hu, H.~Ning, T.~Qiu, Y.~Zhang, and X.~Luo, ``Fog computing based face
  identification and resolution scheme in internet of things,'' \emph{IEEE
  Trans. Ind. Informat.}, vol.~13, no.~4, pp. 1910--1920, Aug 2017.

\bibitem{deloney2008person}
\BIBentryALTinterwordspacing
C.~DeLoney, ``Person identification and gender recognition from footstep sound
  using modulation analysis,'' The Institute for Systems Research, university
  of Maryland, Tech. Rep., 2008. [Online]. Available:
  \url{http://drum.lib.umd.edu/handle/1903/8379}
\BIBentrySTDinterwordspacing

\bibitem{guo_mic_iet}
F.~Guo and X.~Wang, ``Robust footstep identification system based on acoustic
  local features,'' \emph{IET Biometrics}, vol.~6, no.~6, pp. 387--392, 2017.

\bibitem{iot_survey}
L.~D. Xu, W.~He, and S.~Li, ``Internet of things in industries: A survey,''
  \emph{IEEE Trans. Ind. Informat.}, vol.~10, no.~4, pp. 2233--2243, Nov 2014.

\bibitem{bodhi_intruder}
B.~Mukhopadhyay, S.~Anchal, and S.~Kar, ``Detection of an intruder and
  prediction of his state of motion by using seismic sensor,'' \emph{IEEE
  Sensors J.}, vol.~PP, no.~99, pp. 1--1, 2017.

\bibitem{fall_detection_Wang}
C.~Wang, W.~Lu, M.~R. Narayanan, D.~C.~W. Chang, S.~R. Lord, S.~J. Redmond, and
  N.~H. Lovell, ``Low-power fall detector using triaxial accelerometry and
  barometric pressure sensing,'' \emph{IEEE Trans. Ind. Informat.}, vol.~12,
  no.~6, pp. 2302--2311, Dec 2016.

\bibitem{Anch1701:Predicting}
S.~Anchal, B.~Mukhopadhyay, and S.~Kar, ``Predicting gender from footfalls
  using a seismic sensor,'' in \emph{Int. Conf. Commun. Syst. and Netw.
  (COMSNETS)}, Bangalore, India, Jan. 2017.

\bibitem{comp_sen_WSN}
S.~Li, L.~D. Xu, and X.~Wang, ``Compressed sensing signal and data acquisition
  in wireless sensor networks and internet of things,'' \emph{IEEE Trans. Ind.
  Informat.}, vol.~9, no.~4, pp. 2177--2186, Nov 2013.

\bibitem{think_speak}
``{Understand Your Things: The open IoT platform with MATLAB analytics},''
  \url{https://thingspeak.com/}, accessed: 2017-11-10.

\bibitem{Pattern_recognition_bishop}
C.~M. Bishop \emph{et~al.}, \emph{Pattern recognition and machine
  learning}.\hskip 1em plus 0.5em minus 0.4em\relax springer New York, 2006,
  vol.~1.

\end{thebibliography}


\begin{thebibliography}{1}
\providecommand{\url}[1]{#1}
\csname url@samestyle\endcsname
\providecommand{\newblock}{\relax}
\providecommand{\bibinfo}[2]{#2}
\providecommand{\BIBentrySTDinterwordspacing}{\spaceskip=0pt\relax}
\providecommand{\BIBentryALTinterwordstretchfactor}{4}
\providecommand{\BIBentryALTinterwordspacing}{\spaceskip=\fontdimen2\font plus
\BIBentryALTinterwordstretchfactor\fontdimen3\font minus
  \fontdimen4\font\relax}
\providecommand{\BIBforeignlanguage}[2]{{%
\expandafter\ifx\csname l@#1\endcsname\relax
\typeout{** WARNING: IEEEtran.bst: No hyphenation pattern has been}%
\typeout{** loaded for the language `#1'. Using the pattern for}%
\typeout{** the default language instead.}%
\else
\language=\csname l@#1\endcsname
\fi
#2}}
\providecommand{\BIBdecl}{\relax}
\BIBdecl

\bibitem{Pattern_recognition_bishop}
C.~M. Bishop \emph{et~al.}, \emph{Pattern recognition and machine
  learning}.\hskip 1em plus 0.5em minus 0.4em\relax springer New York, 2006,
  vol.~1.

\bibitem{svm}
\BIBentryALTinterwordspacing
C.~Cortes and V.~Vapnik, ``Support-vector networks,'' \emph{Machine Learning},
  vol.~20, no.~3, pp. 273--297, 1995. [Online]. Available:
  \url{http://dx.doi.org/10.1007/BF00994018}
\BIBentrySTDinterwordspacing

\bibitem{ANN}
T.~M. Mitchell, \emph{Artificial neural networks}.\hskip 1em plus 0.5em minus
  0.4em\relax springer, 2006.

\bibitem{lda}
A.~J. Izenman, ``Linear discriminant analysis,'' in \emph{Modern Multivariate
  Statistical Techniques}.\hskip 1em plus 0.5em minus 0.4em\relax Springer,
  2013, pp. 237--280.

\bibitem{faludi2010building}
R.~Faludi, \emph{Building wireless sensor networks: with ZigBee, XBee, arduino,
  and processing}.\hskip 1em plus 0.5em minus 0.4em\relax " O'Reilly Media,
  Inc.", 2010.

\end{thebibliography}

\end{document}


\title{\textbf{\Large{  
			Supplementary File for Person Identification using Seismic Signals generated from Footfalls  }}}
\maketitle

In the supplementary file we present detailed results regarding optimal number of footsteps/samples, comparison among different classifiers (LR, SVM-Lin, SVM-RBF, ANN, and LDA) on footfall dataset, performance related information of the compression techniques, and hardware implementation of the proposed architecture.    

\section{Selection of Optimal Number of footsteps/sample}
\begin{figure}[!h]
\centering
\includegraphics[width=\linewidth]{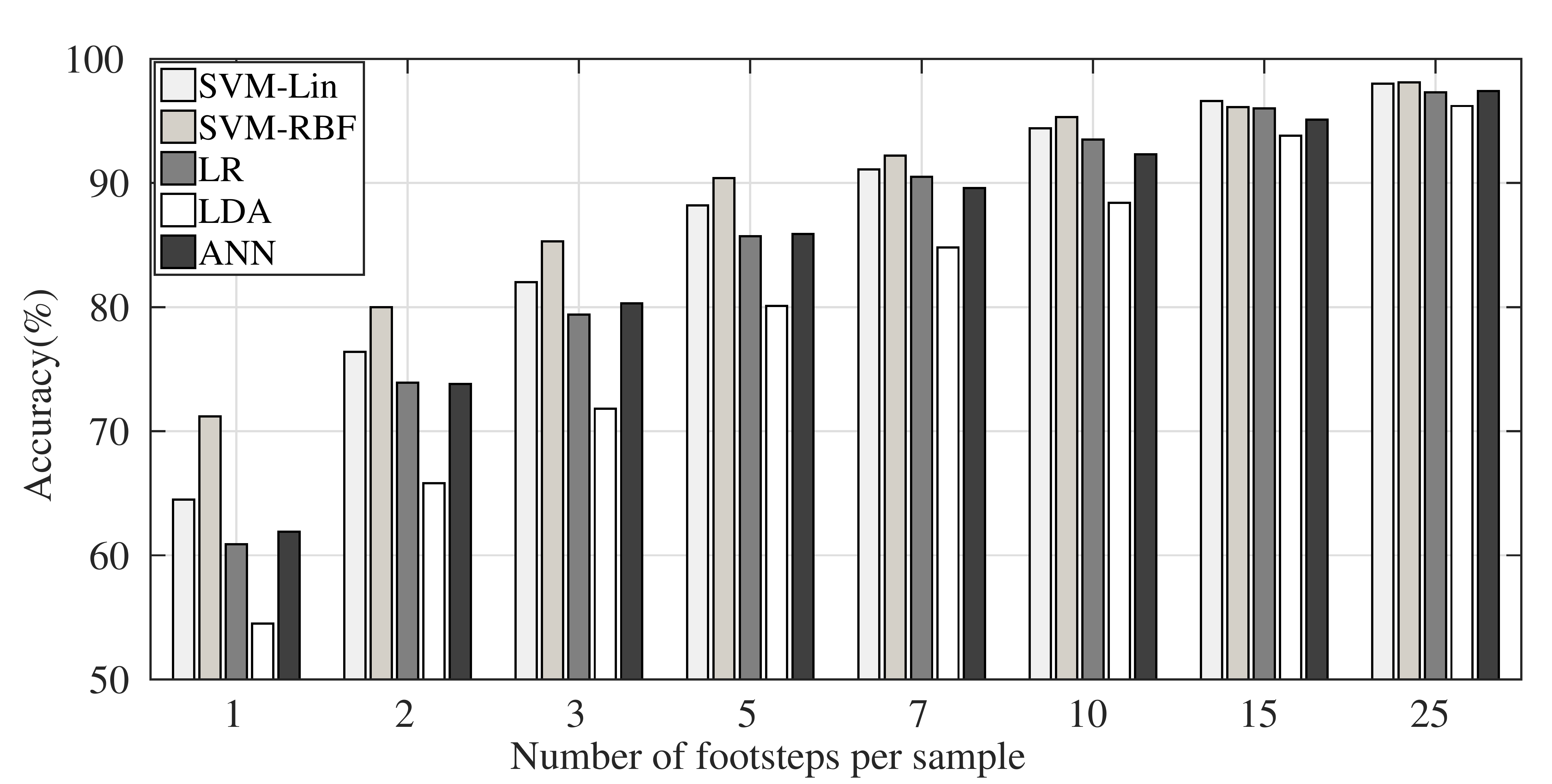}
\caption{Accuracy of different classifier obtained with increasing number of footfalls/sample. (~$f_s$ is 8kHz~)}
\label{fig:acc_footstep_each_clfr}
\end{figure}

Fig.~S\ref{fig:acc_footstep_each_clfr} shows the performance of different classifiers as footsteps/sample are increased from 1 to 25.  It can be clearly observed that as the number of footsteps increases, the performance of linear classifiers (SVM-Lin, LR, LDA) become almost equal to that of non-linear classifiers (SVM-RBF, ANN). The poor performance of the linear classifiers suggests that the classes in the feature space of single footstep/sample are overlapped and are not linearly separable. The overlapping among the classes decreases with the increase in the number of footsteps, as it reduces the effect of noise in the feature space.  

\section{Selection of best suitable Classifiers for person identification}

Table~S\ref{Tb:perf_diff_classifiers} shows the performances (accuracy, precision, recall and F1 Score) of each of the classifiers on the footstep database. It also displays the F1 score of all the individual classes (P1,P2, $\cdots$, P8) and prediction time of a single sample. All  the computations were  performed  using a  64  bit  operating  system running on an Intel(R) Core(TM) i7-4790 (CPU @ 3.60GHz x 8 ) processor with 16.00  GB  RAM. Data set used for performance analysis contained features extracted from 7 consecutive footsteps recorded with a sampling frequency ($f_s$) of 8 kHz.  

\begin{table*}[!t]
\centering
\caption{Performance of different classifiers on the footstep seismic data. (Each sample consisted of 7 footsteps, $f_s$ = 8 kHz)}
\label{Tb:perf_diff_classifiers}
\resizebox{\textwidth}{!}{\begin{tabular}{|c|c|c|c|c|c|c|c|c|c|c|c|c|c|c|}
\hline
\multirow{2}{*}{\textbf{Classifier}} & \multirow{2}{*}{\textbf{Value}} & \multirow{2}{*}{\textbf{Accuracy}} & \multirow{2}{*}{\textbf{Precision}} & \multirow{2}{*}{\textbf{Recall}} & \multirow{2}{*}{\textbf{F1 Score}} & \multicolumn{8}{c|}{\textbf{Individual F1 Scores}} & \textbf{Time(ms)} \\ \cline{7-15} 
 &  &  &  &  &  & \textbf{P1} & \textbf{P2} & \textbf{P3} & \textbf{P4} & \textbf{P5} & \textbf{P6} & \textbf{P7} & \textbf{P8} &  \\ \hline
\textbf{LR} & Mean & 90.13 & 90.01 & 90.04 & 90.09 & 93.70 & 96.15 & 91.12 & 88.86 & 87.51 & 88.80 & 90.51 & 83.46 & 0.1 \\ \hline
\textbf{} & std. & 1.32 & 1.30 & 1.30 & 1.29 & 2.24 & 1.71 & 2.18 & 2.68 & 3.29 & 2.72 & 1.50 & 2.43 &  \\ \hline
\textbf{SVM-Lin} & Mean & 91.90 & 91.82 & 91.81 & 91.89 & 94.94 & 97.59 & 92.58 & 92.22 & 89.46 & 90.67 & 91.91 & 85.23 & 22.6 \\ \hline
\textbf{} & std. & 0.98 & 1& 1& 1.02 & 2.22 & 0.59 & 2.11 & 2.07 & 1.75 & 2.44 & 1.55 & 3.28 &  \\ \hline
\textbf{SVM-RBF} & Mean & 92.29 & 92.19 & 92.20 & 92.25 & 94.72 & 97.82 & 92.36 & 91.57 & 90.20 & 92.52 & 92.68 & 85.66 & 32.8 \\ \hline
\textbf{} & std. & 1.23 & 1.25 & 1.22 & 1.27 & 1.52 & 1.41 & 3.01 & 1.55 & 2.85 & 1.92 & 2.73 & 1.76 &  \\ \hline
\textbf{ANN} & Mean & 88.97 & 88.85 & 88.87 & 88.97 & 91.38 & 95.43 & 90.21 & 87.53 & 87.31 & 88.21 & 90.65 & 80.12 & 1.7 \\ \hline
\textbf{} & std. & 1.30 & 1.32 & 1.31 & 1.28 & 1.49 & 0.57 & 3.36 & 2.14 & 3.97 & 3.30 & 2.96 & 4.00 &  \\ \hline
\textbf{LDA} & Mean & 84.79 & 84.81 & 84.77 & 85.23 & 86.34 & 95.19 & 86.91 & 78.38 & 84.33 & 85.64 & 87.32 & 74.37 & 2.5 \\ \hline
 & std. & 1.17 & 1.18 & 1.20 & 1.15 & 1.46 & 0.52 & 3.05 & 2.26 & 2.06 & 2.64 & 1.93 & 3.01 &  \\ \hline
\end{tabular}}
\end{table*}


LR~\cite{Pattern_recognition_bishop} is a binary classifier with categorical dependent variable i.e. class 1 and class 2 are represented by 0 and 1. The function used by LR for prediction is:
\begin{equation}
\label{Eq:LR}
 f_{LR}(\boldsymbol{x};\boldsymbol{\theta})=\frac{1}{1+ e^{-\boldsymbol{\theta^T} \boldsymbol{x}} }
\end{equation}
where  $\boldsymbol{\theta} (\in \mathfrak{R}^{n+1})$
 is the model parameter or weights, $\boldsymbol{x}\in\mathfrak{R}^{n+1} \ ([x_0=1, \ x_1, \ \cdots \ x_n ]^T) $ is the feature vector corresponding to a single sample, and $n$ is the number of features. A test sample $\boldsymbol{x_{test}}$ belongs to class 1 if $f_{LR}(\boldsymbol{x_{test}};\boldsymbol{\theta}) > 0.5 $ and to class 0 if $f_{LR}(\boldsymbol{x_{test}};\boldsymbol{\theta}) < 0.5 $. Multi-class classification by LR was implemented by using one-vs.-rest technique. The simplicity of the LR model makes it easily implementable in a lower end embedded processors. However, its biggest drawback is its performance with data having lower number of footsteps per sample (60\,\% in case of single footstep/sample, refer Fig.~S\ref{fig:acc_footstep_each_clfr}).

SVM-Lin~\cite{svm} showed almost similar prediction accuracy with respect to LR, as both of them linearly separates the dataset. However here the classes are labelled as +1 (class 1) and -1 (class 2). It is modelled using the function 
\begin{equation}
\label{Eq:SVM-lin}
 f_{SVM-Lin}(\boldsymbol{x};\{ \boldsymbol{w},b\})= sign(  \boldsymbol{w}^T\boldsymbol{x} + b )
\end{equation}
where  $\boldsymbol{w}(\in \mathfrak{R}^{n})$ is the weight vector, $\boldsymbol{x}$ is the feature vector and $b \in \mathfrak{R}$ is a scalar. 
A grid-search was performed to find the optimal value of the hyper parameter $C (\in \mathfrak{R})$ and was set to 1. From Fig.~S\ref{fig:acc_footstep_each_clfr}, it can be seen that SVM-Lin outperforms LR in terms of accuracy by 2-3\,\% when number of footsteps/sample is below 7. It is due to the soft margin of SVM-Lin, as it allows some miss-classification in the training dataset during model training.

SVM is well known for its Kernel trick which works very efficiently when the classes in the feature space are not linearly separable. SVM-RBF is modeled by the function
\begin{equation}
\label{Eq:SVM-RBF}
f_{SVM-RBF}(\boldsymbol{x};\{ \alpha_i,K,\boldsymbol{x_{sv}},b\})= sign (\sum_{i = 1}^{m} \alpha_i K(\boldsymbol{x_{sv}^{(i)}},\boldsymbol{x})+ b )
\end{equation}
where  $K$ is the Kernel function, $\boldsymbol{x_{sv}^{(i)}}$ is the $i^{th}$ support vector, $\alpha_i$ is a positive constant, $\boldsymbol{x}$ is the feature vector of the test sample, $b \in \mathfrak{R}$ is a scalar and $m$ is the number of support vectors. The Kernel function for RBF is 
\begin{equation}
\label{Eq:kernel}
K (\boldsymbol{x_{sv}^{(i)}},\boldsymbol{x}) = exp(-\gamma ||\boldsymbol{x_{sv}^{(i)}} - \boldsymbol{x}||^2)
\end{equation}
Soft-margin SVM-RBF has two hyper parameters $C$ and $\gamma$. The values of C and $\gamma$ were obtained using a grid search and were found to be 100 and 0.001. SVM-RBF performed consistently well among the other classifiers because, its Kernel($K$) transformed the non linearly separable low dimension features to a linearly separable high dimensional feature. Table~S\ref{Tb:conf_svm-rbf_25footstep} shows the confusion matrix of SVM-RBF on a dataset obtained using features from 25 consecutive footfalls. 

ANN~\cite{ANN} (single hidden layer) was the most computationally expensive learning technique used in this paper. It took maximum amount of training time w.r.t to the other classifiers. However, its prediction time was better than SVM-Lin and SVM-RBF. The number of neural net used in the single layer ANN was 40 (obtained by grid search). Performance-wise it does not produce better results than other classifiers. The performance of LDA~\cite{lda} was worst among all the classifiers. It is because LDA assumes the features are drawn from a Gaussian distribution and they share a common covariance matrix.  


\begin{table}[!h]
\caption{Confusion Matrix of SVM-RBF for 25 footsteps/sample and 500 Hz sampling frequency}
\label{Tb:conf_svm-rbf_25footstep}
\noindent\begin{tabular}{cc*{8}{|E}|}
\cline{3-10}
\multirow{8}{*}{\rotatebox[origin=c]{90}{\textbf{Predicted Class}}}
&\textbf{P1} & 1 & 0 & 0 & 0.01 & 0 & 0 & 0 & 0 \\ \cline{3-10}
&\textbf{P2} & 0 & 1& 0 & 0 & 0 & 0 & 0 & 0  \\ \cline{3-10}
&\textbf{P3} & 0 & 0 & 0.96 & 0 & 0 & 0 & 0 & 0  \\ \cline{3-10}
&\textbf{P4} & 0 & 0 & 0.02 & 0.98 & 0 & 0 & 0 & 0  \\ \cline{3-10}
&\textbf{P5} & 0 & 0 & 0 & 0 & 1 & 0 & 0.02 & 0.02 \\ \cline{3-10}
&\textbf{P6} & 0 & 0 & 0 & 0.01 & 0 & 1& 0 & 0  \\ \cline{3-10}
&\textbf{P7} & 0 & 0 & 0 & 0 & 0 & 0 & 0.96 & 0.02  \\ \cline{3-10}
&\textbf{P8} & 0 & 0 & 0.02 & 0 & 0 & 0 & 0.02 & 0.96 \\ \cline{3-10}
& \multicolumn{1}{c}{} & \multicolumn{1}{c}{\textbf{P1}} & \multicolumn{1}{c}{\textbf{P2}} 
 & \multicolumn{1}{c}{\textbf{P3}} & \multicolumn{1}{c}{\textbf{P4}} & \multicolumn{1}{c}{\textbf{P5}} 
 & \multicolumn{1}{c}{\textbf{P6}} & \multicolumn{1}{c}{\textbf{P7}} & \multicolumn{1}{c}{\textbf{P8}}\\ 
& \multicolumn{1}{c}{} &    \multicolumn{8}{c}{\textbf{Actual Class}} 
\end{tabular}
\end{table}

\begin{table*}[]
\centering
\caption{Prediction accuracy of classifiers with and without compression technique on the footfall event. (NC: Not compressed ($f_s$=8 kHz))}
\label{Tb:comp_ds8bp_ds16}
\begin{tabular}{|c|c|c|l|c|c|c|}
\hline
\multirow{2}{*}{\textbf{Classifiers}} & \multirow{2}{*}{\textbf{Compression}} & \multicolumn{5}{c|}{\textbf{Number of Footsteps/sample}} \\ \cline{3-7} 
 &  & \textbf{1} & \multicolumn{1}{c|}{\textbf{2}} & \textbf{3} & \textbf{5} & \textbf{7} \\ \hline
\multirow{3}{*}{\textbf{\begin{tabular}[c]{@{}c@{}}SVM\\ RBF\end{tabular}}} & \textbf{NC} & 71.20 & 80.00 & 85.3 & 90.40 & 92.20 \\ \cline{2-7} 
 & \textbf{DS16} & 70.49 & 80.15 & 84.69 & 89.69 & 91.33 \\ \cline{2-7} 
 & \textbf{DS8BP} & 73.37 & 81.22 & 85.51 & 90.16 & 93.10 \\ \hline
\multirow{3}{*}{\textbf{\begin{tabular}[c]{@{}c@{}}SVM\\ Linear\end{tabular}}} & \textbf{NC} & 64.50 & 76.40 & 82.00 & 88.20 & 91.10 \\ \cline{2-7} 
 & \textbf{DS16} & 64.03 & 76.11 & 81.46 & 87.17 & 90.36 \\ \cline{2-7} 
 & \textbf{DS8BP} & 67.51 & 78.15 & 83.85 & 89.08 & 92.63 \\ \hline
\multirow{3}{*}{\textbf{\begin{tabular}[c]{@{}c@{}}Logistic\\ Regression\end{tabular}}} & \textbf{NC} & 60.90 & 73.90 & 79.40 & 85.70 & 90.50 \\ \cline{2-7} 
 & \textbf{DS16} & 60.62 & 73.10 & 78.59 & 85.77 & 89.34 \\ \cline{2-7} 
 & \textbf{DS8BP} & 64.92 & 75.50 & 80.16 & 85.77 & 91.40 \\ \hline
\end{tabular}
\end{table*}

 \section{Performance   of   the   proposed   Compression   and   De-compression Technique}

\begin{figure}[]
\centering
\includegraphics[width=\linewidth]{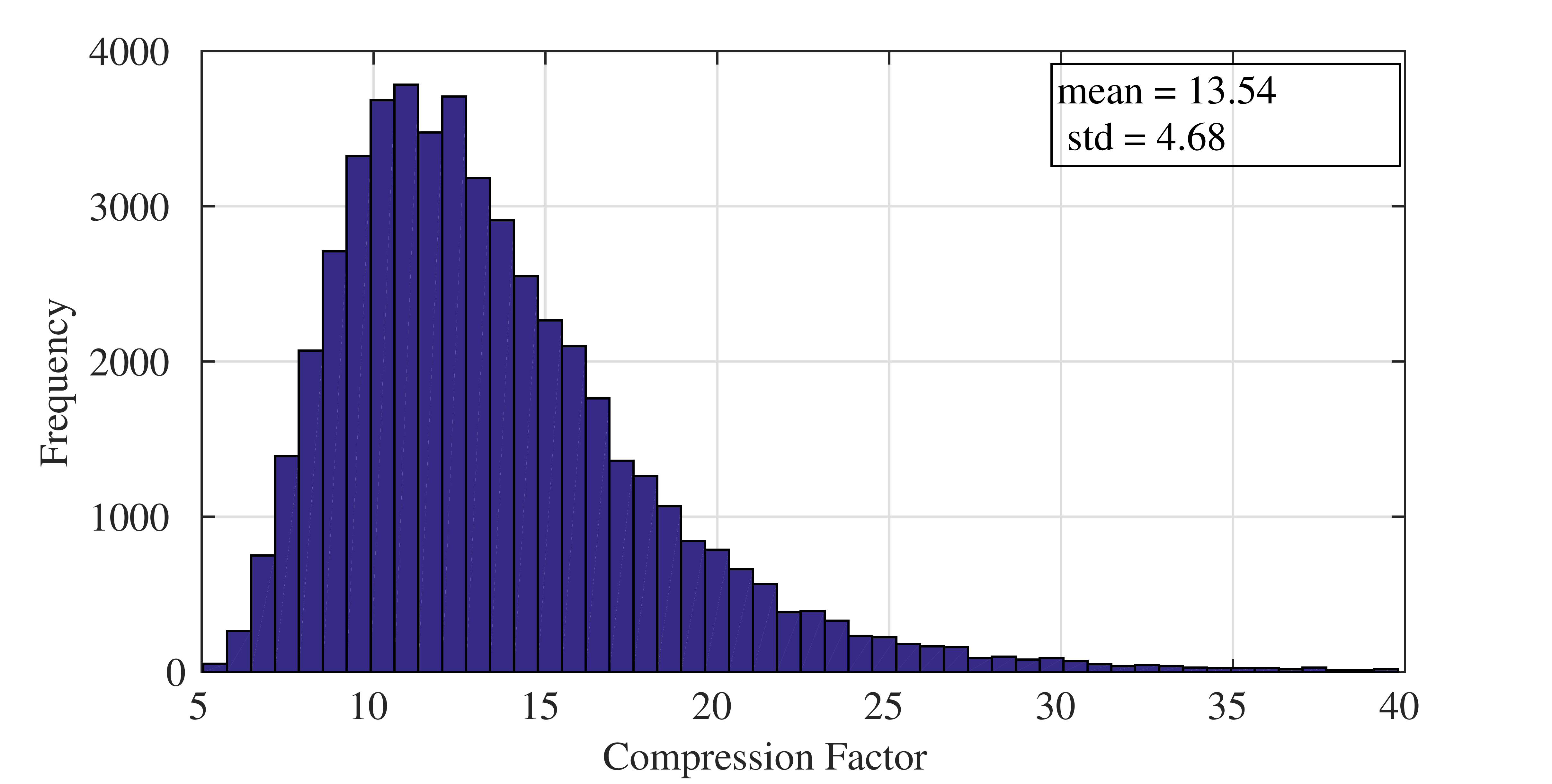}
\caption{Histogram of Compression Factor [length($Sig_{comp}$)/length($ Sig_{ds}$] achieved by \textit{DS8BP}. }
\label{fig:compression_hist}
\end{figure}

Fig.S\ref{fig:compression_hist} shows the histogram of compression factors i.e. length(compressed signal-$Sig_{comp}$)/length(downsampled signal-$ Sig_{ds}$) achieved by \textit{DS8BP}. The histogram includes all the footfall signals (Table III of the main manuscript) whose compression factor is between 5($L_{GC}$) and 40($H_{GC}$). The average compression factor achieved by \textit{DS8BP} is 13.54 with a standard deviation of 4.68.

\begin{figure}[]
\centering
\includegraphics[width=\linewidth]{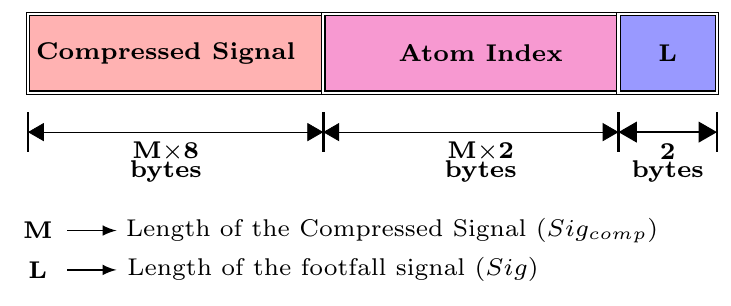}
\caption{Datagram representation of a single footstep. $10 \times M+2$~bytes are required to represent a single footstep.}
\label{fig:datagram}
\end{figure}

\begin{figure}[]
\centering
\includegraphics[width=.9\linewidth]{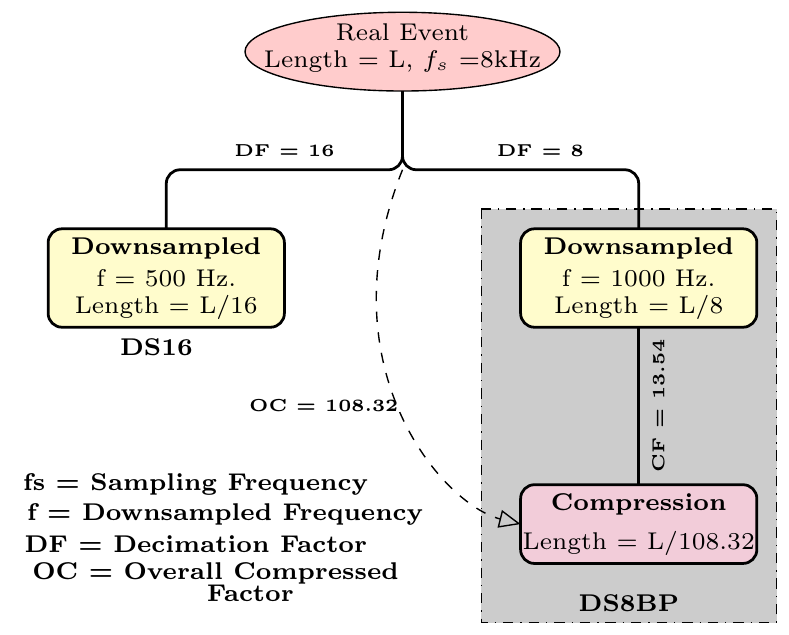}
\caption{Comparison between \textit{DS8BP} and \textit{DS16} regarding comprehensibility capability.  }
\label{fig:compression_sys}
\end{figure}

The datagram used for transmitting footfall signal from $EP$ to $EP^{++}$ over a Zigbee network is shown in Fig.~S\ref{fig:datagram}. Fig.~S\ref{fig:compression_sys} draws a graphical comparison between \textit{DS16} and \textit{DS8BP}.


Table.~S\ref{Tb:comp_ds8bp_ds16} shows the prediction accuracy of SVM-RBF, SVM-Lin, and LR when compression techniques \textit{DS8BP} and \textit{DS16} are used on the original footfall signal. In case of \textit{DS8BP}, the features are extracted from the de-compressed signal ($Sig_{rec}$, Algorithm 2 in the main manuscript). And in \textit{DS16} features are extracted from the downsampled (decimation factor = 16) footfall signals. In NC (No Compression) features are extracted from the original 8 kHz signal. Table.~S\ref{Tb:comp_ds8bp_ds16} clearly shows that the performance of NC and \textit{DS16} is almost similar. However, \textit{DS8BP} increases classifiers accuracy (2\%-4\%), (1\%-2\%)  in case of single footstep/sample and 7 footsteps/sample when compared to other techniques.

\begin{figure}[]
\centering
       \begin{subfigure}[b]{.5\textwidth}
       		  \includegraphics[width=.9\linewidth]{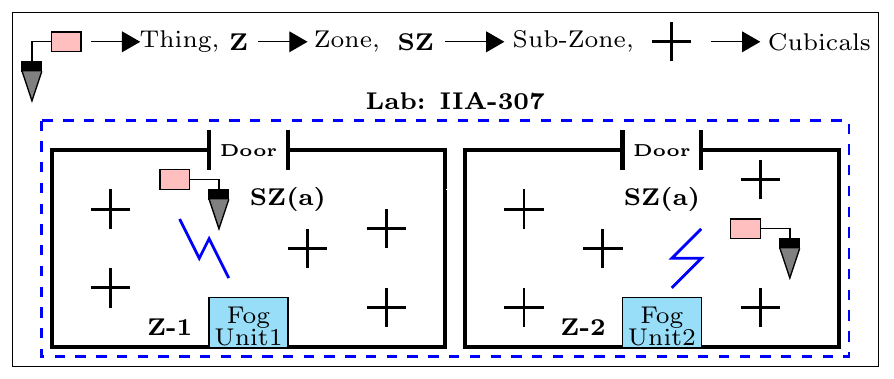}	
                \caption{Floor plan for implementing human identification system using seismic sensor.   }
                \label{fig:PrototypeFog}
        \end{subfigure}%
        
        \begin{subfigure}[b]{.45\textwidth}
                \includegraphics[width=\linewidth]{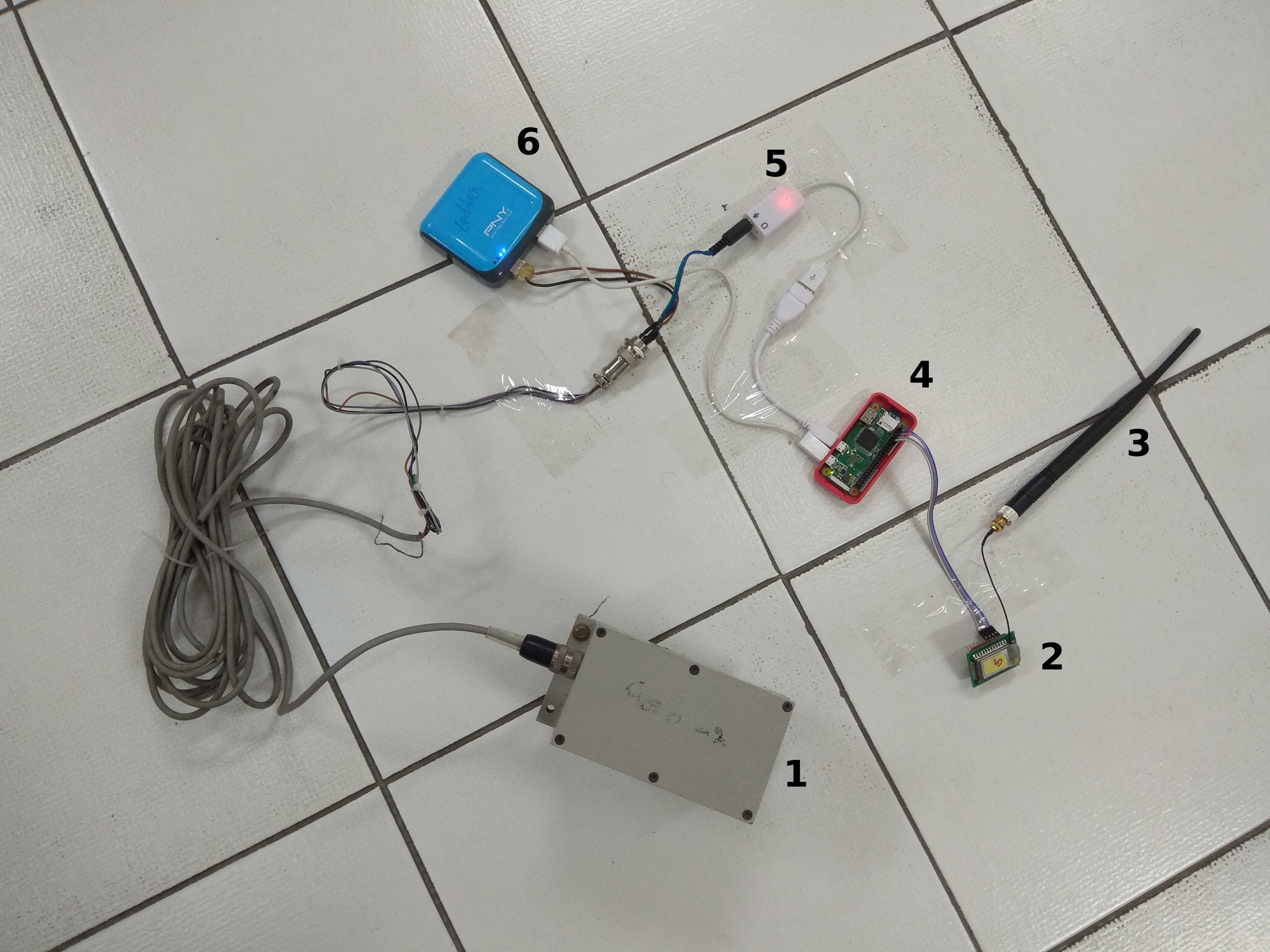}
                \caption{A \textit{Thing} constituted the following components ; 1)~Geophone; 2)~Xbee 868 LP; 3)~Antennae; 4)~Raspberry Pi Zero ($EP$); 5)~Sound Card (ADC); 6)~Battery.  }
                \label{fig:EP}
        \end{subfigure}%
        
        \begin{subfigure}[b]{0.35\textwidth}
        \centering
                \includegraphics[width=\linewidth,]{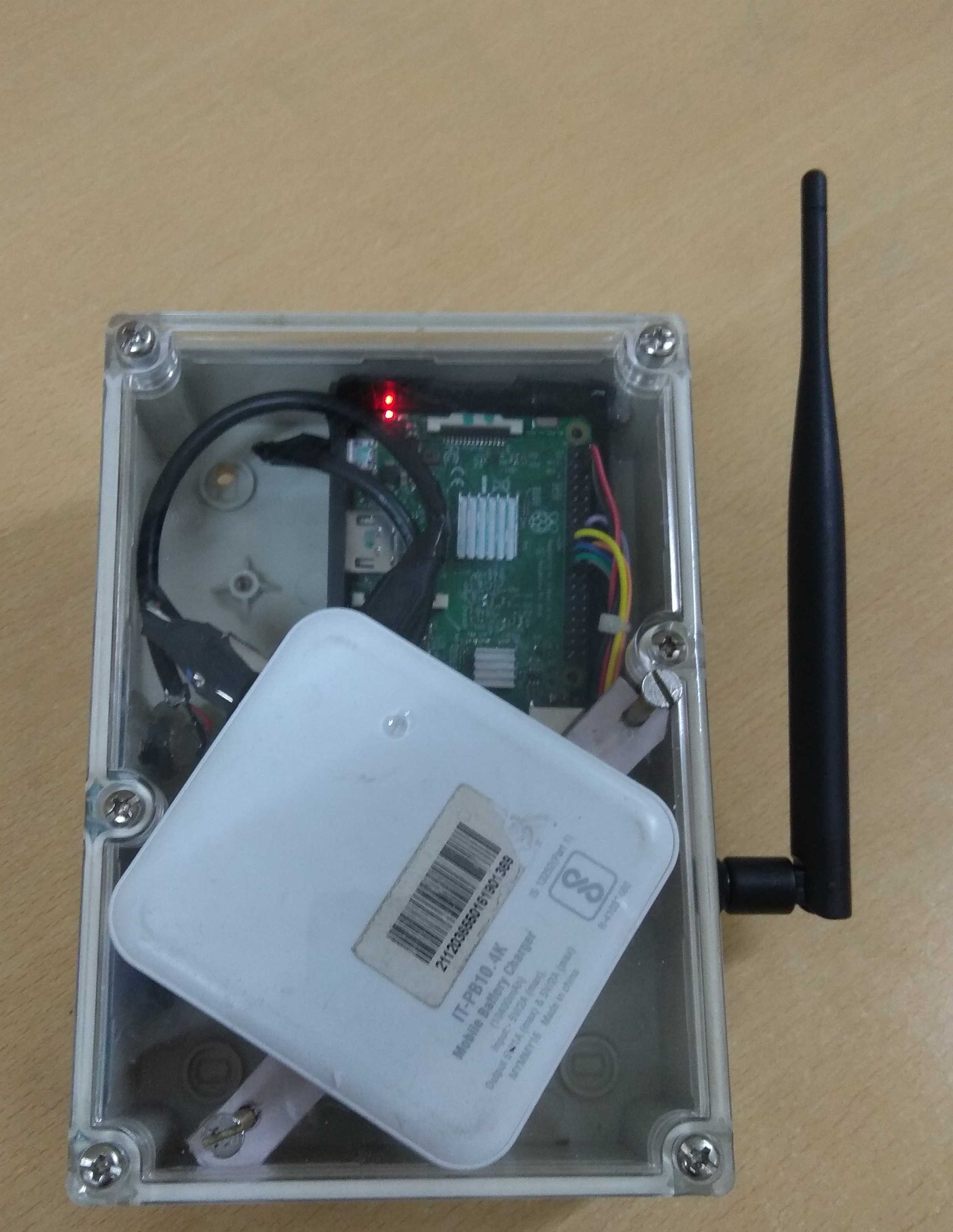}
                \caption{A \textit{Fog} unit consisted of the following components: Raspberry Pi 3 model B ($EP^{++}$); Xbee 868 LP; Battery.   }
                \label{fig:EP++}
        \end{subfigure}%
        \caption{Floor planning and hardware components for implementing person identification system.}\label{fig:hardare_comp}
\end{figure}

\section{Hardware Implementation}

\begin{figure}[!t]
\centering
\includegraphics[width=\linewidth]{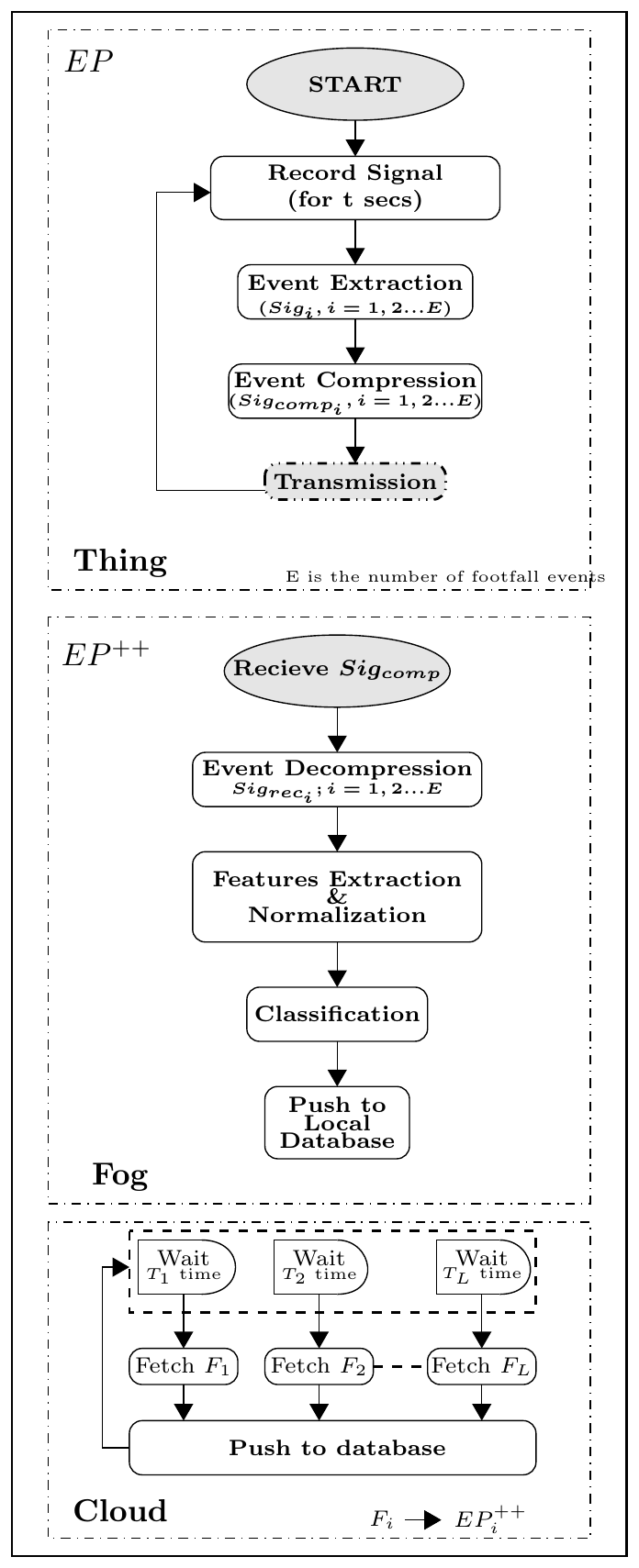}
\caption{Flow charts of Algorithms running in \textit{$EP$}, \textit{$EP^{++}$} and \textit{Cloud}.}
\label{fig:flowchart}
\end{figure}

\begin{figure}[]
\centering
        \begin{subfigure}[b]{.5\textwidth}
                \includegraphics[width=\linewidth]{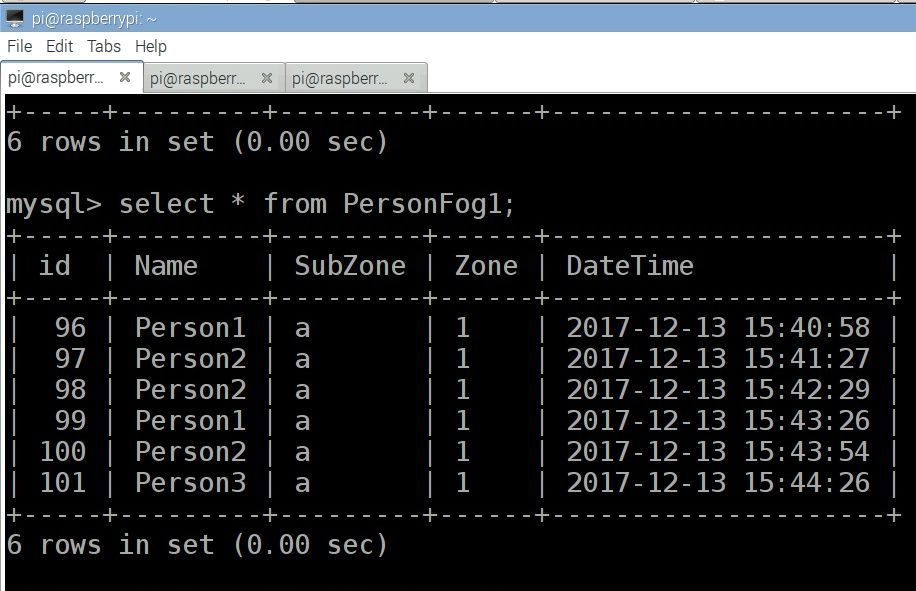}
                \caption{Local database (PersonFog1) hosted in $EP^{++}$ of \textit{Fog Unit 1}. }
                \label{fig:table_fog1}
        \end{subfigure}%
        
        \begin{subfigure}[b]{0.5\textwidth}
        \centering
                \includegraphics[width=\linewidth,]{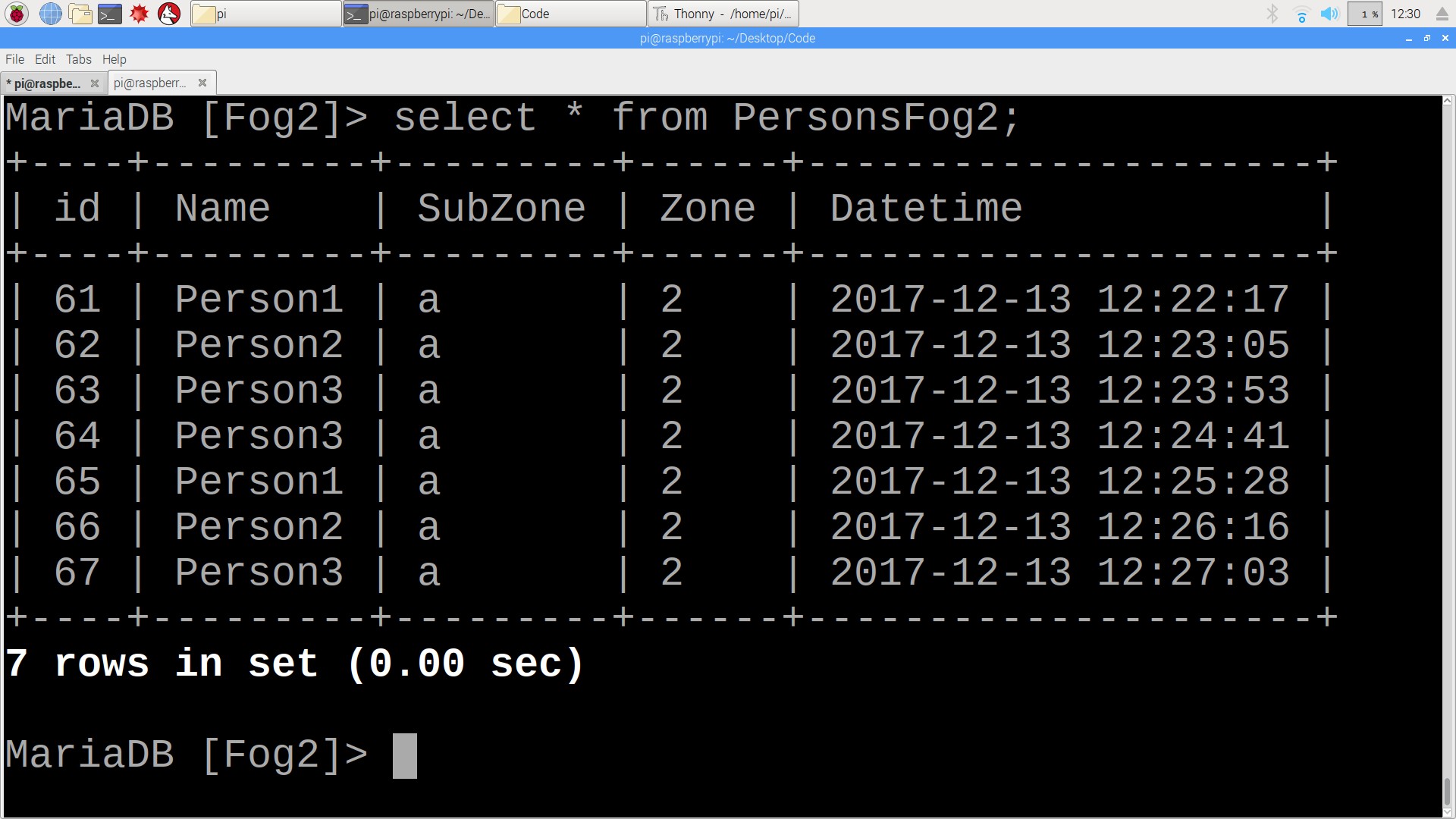}
                \caption{Local database (PersonsFog2) hosted in $EP^{++}$ of \textit{Fog Unit 2}. }
                \label{fig:table_fog2}
        \end{subfigure}%
        
            \begin{subfigure}[b]{0.5\textwidth}
        \centering
                \includegraphics[width=\linewidth,]{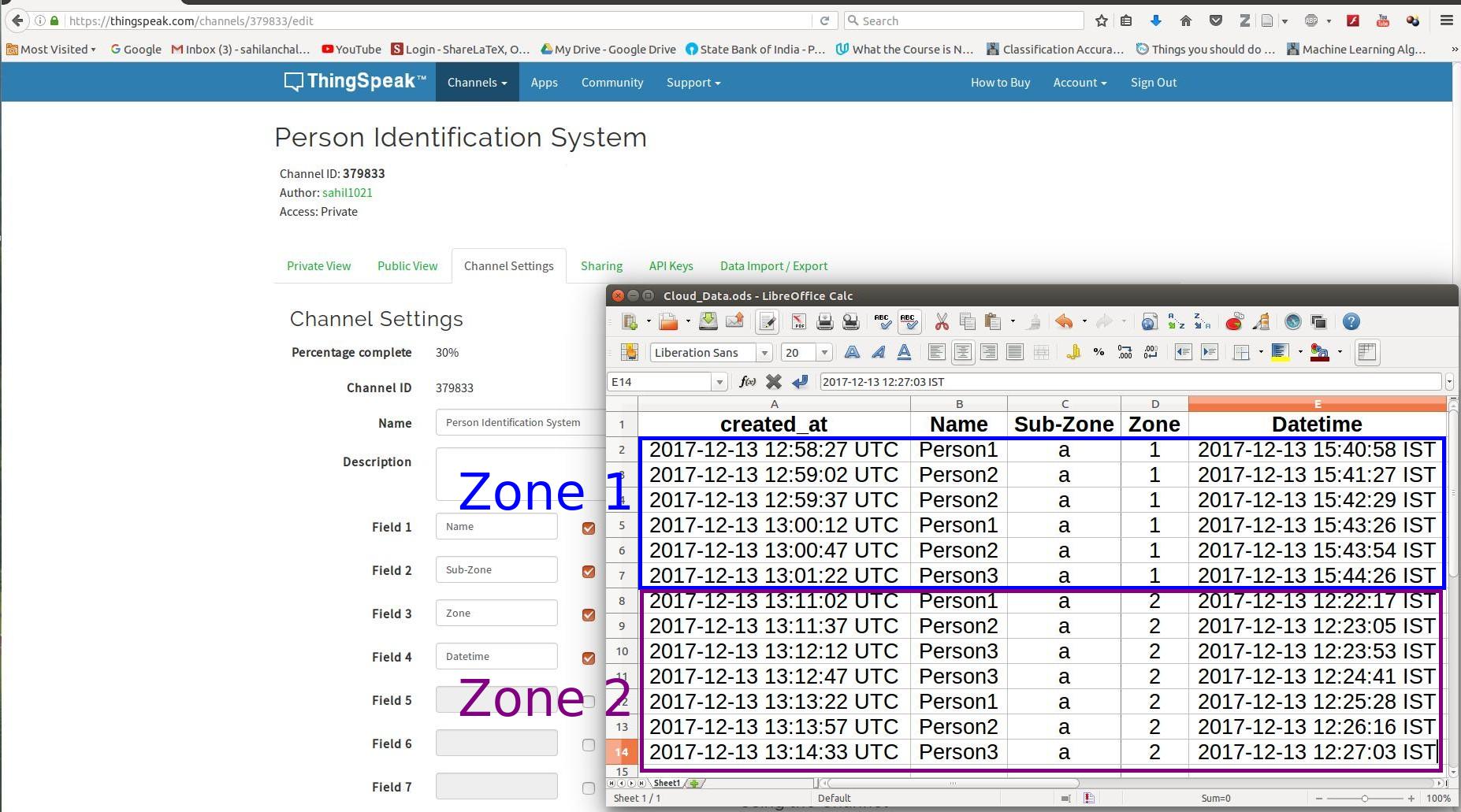}
                \caption{Global dtabase in the \textit{Cloud}.  }
                \label{fig:table_cloud}
        \end{subfigure}%
        \caption{Local databases in the \textit{Fog} and global database in the \textit{Cloud}.}\label{fig:tables_fog_cloud}
\end{figure}

We have implemented \textit{Fog} computing based human identification system in our lab using geophones. Fig.~S\ref{fig:PrototypeFog} shows the floor plan of our implementation. We have divided the lab (IIA-307) into two zones: Z-1 and Z-2. Each of the zones consisted only a single Sub-Zone (SZ(a)) and each Sub-Zone had only one \textit{Thing}. Fig.~S\ref{fig:EP} and Fig.~S\ref{fig:EP++} show components used for implementing a \textit{Things} and a \textit{Fog Unit}. Fig.~S\ref{fig:flowchart} shows the flowcharts of algorithms running on the processors present in \textit{Things}, \textit{Fog}, and \textit{Cloud}. $EP$ records a $t$ sec signal and extracts footfall events from it. It then compresses the footfall events individually using \textit{DS8BP} and transmit them to $EP^{++}$ using XBee~\cite{faludi2010building}. The value of $t$ depends on the application for which the system is being used. Here, we have set the value of $t$ to 10 sec. The processor in the \textit{Fog} ($EP^{++}$) extracts features from the footfall events and then classifies the signal. It finally stores the predicted results in its local database. The \textit{Cloud} fetches information from these local databases (present in the $EP^{++}$s) periodically and updates its own database. $T_j$ ($j=1,...,L$) is an application specific variable, it determines the frequency by which the \textit{Cloud} updates its tables. Here $F_i$ represents $i^{th}$ \textit{Fog Unit} (in our experiment i={1,2}). Six students had volunteered (three in each zone) for our experiment on human identification. \footnote{To prevent identity disclosure the students were denoted as Person\textit{i} (where \textit{i}=1,2,3).} 
Fig.S\ref{fig:table_fog1} and Fig.S\ref{fig:table_fog2} displays the local databases present with in the two \textit{Fog Unit}s. Fig.S\ref{fig:table_cloud} shows the global database present in the \textit{Cloud} which periodically copies the database of the \textit{Fog Units}. 

 \bibliographystyle{IEEEtran}
 \bibliography{reference}